\documentclass[12pt]{article}
\pdfoutput=1 
\usepackage{amsmath}
\usepackage{graphicx,psfrag,epsf}
\usepackage{enumerate}
\usepackage{natbib}
\usepackage{amsthm} 
\usepackage{amssymb}
\usepackage{subcaption}
\usepackage{url} % not crucial - just used below for the URL 
\newtheorem{theorem}{Theorem}

\newcommand{\blind}{0}

\begin{document}

%\bibliographystyle{natbib}

%%%%%%%%%%%%%%%%%%%%%%%%%%%%%%%%%%%%%%%%%%%%%%%%%%%%%%%%%%%%%%%%%%%%%%%%%%%%%%

\if0\blind
{
  \title{\bf Functional Gaussian Process \\ for Large Scale Bayesian Nonparametric Analysis}
  \author{Leo L. Duan\\
    University of Cincinnati\\
    and\\
           Xia Wang\\
    University of Cincinnati\\
        and\\
    Rhonda D. Szczesniak  \thanks{
    	Corresponding author. Address: 3333 Burnet Ave, MLC 5041, Cincinnati, OH 45229. Phone:(513)803-0563, email: rhonda.szczesniak@cchmc.org. The authors gratefully acknowledge the Cystic Fibrosis Foundation Research and Development Program (grant number R457-CR11) for the support of this research.
    	}\\
    Cincinnati Children's Hospital Medical Center}
  \maketitle
 \fi

\if1\blind
{
  \bigskip
  \bigskip
  \bigskip
  \begin{center}
    {\LARGE\bf Functional Gaussian Process Model\\ for Bayesian Nonparametric Analysis}
\end{center}
  \medskip
} \fi

%\bigskip
\begin{abstract}

Gaussian process is a theoretically appealing model for nonparametric analysis, but its computational cumbersomeness hinders its use in large scale and the existing reduced-rank solutions are usually heuristic. In this work, we propose a novel construction of Gaussian process as a projection from fixed discrete frequencies to any continuous location. This leads to a valid stochastic process that has a theoretic support with the reduced rank in the spectral density, as well as a high-speed computing algorithm. Our method provides accurate estimates for the covariance parameters and concise form of predictive distribution for spatial prediction. For non-stationary data, we adopt the mixture framework with a customized spectral dependency structure. This enables clustering based on local stationarity, while maintains the joint Gaussianness. Our work is directly applicable in solving some of the challenges in the spatial data, such as large scale computation, anisotropic covariance, spatio-temporal modeling, etc. We illustrate the uses of the model via simulations and an application on a massive dataset.
\end{abstract}

\noindent%
{\it Keywords:} {Bayesian Sampling}, {Non-stationary Gaussian},{ Spectral Projection}, {Sparse Spectral Density}, {Uncountable Location Set} 
\vfill

\newpage

\section{Introduction}

Gaussian process has a wide range of applications  from computer experiment emulations \citep{loeppky2009choosing} to spatial data analysis \citep{cressie2015statistics}. The covariance is the crucial component but it involves two crucial challenges: its evaluation in the likelihood is prohibitively cumbersome and the consideration for non-stationarity is difficult.

Firstly, to address the computational issue, various reduced-rank approaches have been proposed: Nystr{\"o}m method uses truncation to the top $m$ eigenvectors in the matrix \citep{smola2000sparse}; the Gaussian predictive process \citep{banerjee2008gaussian} models the full observation as the prediction mean from a small set  (with size $m$) of the knot process; the spatial random effects model \citep{cressie2008fixed} treats the covariance as a quadratic transform of $m$ predetermined basis functions. With small $m$, these methods reduces the computational burden and makes the Gaussian process fitting viable in large data. However, the small $m$ usually incurs sacrifices in resolution and its choice is commonly heuristic.

In spectral domain, some alternative approaches have been proposed to overcome the computational obstacle. This pioneering work was Whittle's likelihood \citep{whittle1953analysis} in time series analysis, where the covariance is approximated by a discrete Fourier transform of the spectral density, if the data are regularly spaced. Multiple research methods have been proposed to accommodate the irregularity in the lattice: \cite{fuentes2007approximate} uses lattice binning and mean filling to minimize the effects of irregularity; \cite{stroud2014bayesian} uses lattice embedding and Bayesian latent framework to estimate the incomplete lattice data; \cite{xu2015bayesian} models the randomly located data as realization of a Gaussian Markov random field conditional on the lattice. These methods have very appealing computational advantage as the likelihood evaluation is $O(n\log_2n)$ (faster than the reduced rank methods) and do not involve resolution reduction. On the other hand, there is not much theoretic development on the relaxed assumption about the lattice. Since conditioning on any arbitrary points off the data lattice would involve changes in the data lattice and the matrix decomposition, hence the likelihood does not lead to a valid stochastic process in the continuous data space; likewise, prediction formulation (Kriging \citep{cressie2015statistics}) is much restrictive as it only projects to the lattice points.

Secondly, non-stationarity is very common in real-world data collected over large space. A general strategy is letting the covariance function (or the spectral density) vary with locations, as has been studied by \cite{paciorek2006spatial} and \cite{ anderes2011local}. In this regard, \cite{priestley1965evolutionary} proposed the idea of ``semi-stationary process'', which matured to ``locally stationary process''  \citep{dahlhaus2000likelihood} with efficient algorithms \citep{guinness2013transformation, guinness2015likelihood} to partition the full domain into small stationary regions. For this spatial partitioning, an infinite mixture process provides more theoretically sound support from a probabilistic point of view: given the latent class assignment (also known as ``clustering''), the local stationarity is achieved inside each class. The work in this area includes a spatially varying weight and stationary Gaussian process component \citep{duan2007generalized,rodriguez2010latent, rodriguez2011nonparametric}.  Nevertheless, it is important to point out the discrepancy between the locally stationary process and the mixture construction: in the former, the whole random vector is still correlated across different stationary regions; in the latter, it is assumed independent given the different latent class assignment (i.e. after the data are clustered). In the mixture distribution case, to represent the correlation on the full domain, one has to use the marginal distribution, which is no longer Gaussian.

In this paper, we propose a new spectral construction of the Gaussian process. Instead of constraining the data to a lattice, we only fix the frequencies to a finite support set. This allow us to construct a valid Gaussian process as a noisy projection from a finite spectral set to any arbitrary and uncountable set in $d$-dimensional data space $\mathbb{R}^d$. The duality in frequency lattice and continuous space enables us to take advantages in both the fast sampling algorithm and the common tools such as Kriging. Due to the sparsity in the frequency support, the rank of the matrix can be automatically reduced and the computation is further accelerated to $O(m\log_2n)$. As for the non-stationary consideration, using different way of projection from the shared frequency random vector  leads to a valid non-stationary Gaussian covariance. We customize the mixture framework with this spectral dependency and show that the distribution is still multivariate Gaussian, even conditional on the different cluster labeling. We illustrate the advantages of the proposed work with simulations and a large spatio-temporal application with 1,343,752 data entries.

\section{Functional Gaussian Process}

\subsection{Spectral Construction}

As shown by \cite{priestley1965evolutionary} and later by \cite{higdon1998process}, a weakly stationary Gaussian process can be represented as $\tilde{Z}_{\bf s}=\int_{\omega\in\mathbb{R}^d} exp(i {\bf s}^T{\boldsymbol \omega}) g^{1/2}(\boldsymbol\omega) Y({\boldsymbol \omega})d {\boldsymbol \omega} +\xi_{\bf s}$. In this formulation, $Y (\boldsymbol{\omega}) \sim CN(\bf 0, I)$ is an orthogonal complex normal process folded over $\bf 0$, that is, $Y (\boldsymbol{\omega})= Y_1(\boldsymbol{\omega}) + i Y_2(\boldsymbol{\omega})$ with $Y_1(\boldsymbol{\omega}), Y_2(\boldsymbol{\omega}) \stackrel{indep}{\sim} N (\bf 0, I)$ with constraint that $Y(\omega)=Y(-\omega)$. We added the remaining $\xi_{\bf s}\stackrel{iid}{\sim}N(0,\sigma^2)$ to ensure the full rank in the finite dimensional density. By Bochner's theorem, The function $g(\boldsymbol\omega) = (g^{1/2}(\boldsymbol\omega))^2$ is a positive and real value function, which can be represented as the forward Fourier transform of the covariance function.

We now give a similar construction but using discrete representation:

\begin{equation}
Z_{\bf s}=\sum_{\{\boldsymbol\omega_l\}_l=1...m} exp(i {\bf s}^T{\boldsymbol \omega_l}) g^{1/2}(\boldsymbol\omega_l) Y({\boldsymbol \omega_l})/\sqrt{n} +\xi_{\bf s}
\label{stationary_construction}
\end{equation}
where each $\boldsymbol{\omega}_l$ represents a $d$-dimensional coordinate from a Cartesian product set of $\{-\frac{m_1/2}{n_1/2}\Delta_1,-\frac{m_1/2-1}{n_1/2}\Delta_1,...,\frac{m_1/2}{n_1/2}\Delta_1\} \times ... \times \{-\frac{m_d/2}{n_d/2}\Delta_d,-\frac{m_d/2-1}{n_d/2}\Delta_d,...,\frac{m_d/2}{n_d/2}\Delta_d\}$ where $m_{(.)} \le n_{(.)}$, $m=m_1m_2...m_d$ is the total number of coordinates. They have fixed increment and symmetry about $\bf 0$. We assume the mild condition that $g(\boldsymbol\omega)\approx 0$ if $\boldsymbol\omega$ falls outside of the region $\mathbb{W}=(-\frac{m_1}{n_1}\Delta_1,\frac{m_1}{n_1}\Delta_1)\times...\times(-\frac{m_d}{n_d}\Delta_d,\frac{m_d}{n_d}\Delta_d)$. Formally, in each sub-dimension of $\boldsymbol\omega$,  for any arbitrarily small $\epsilon>0$ there is an $\boldsymbol\omega_0(\epsilon)$ such that if $|\boldsymbol\omega|>|\boldsymbol\omega_0(\epsilon)|$, $g(\boldsymbol\omega)<\epsilon$. This condition is satisfied by most of the spectral density functions, since the detectable frequencies is always bounded by the sampling rate (e.g. the minimum distance). And commonly used covariance family, such as Mat\'ern, has the spectral density $g(\rho, \omega)$ as a decreasing function in both the range parameter $\rho$ and frequency $\omega$: when $\rho$ is large, $g$ declines rapidly in $ \omega$; when $\rho$ is small, we can scale up the location unit so that $\rho$ increases accordingly; when $\rho$ is extremely small, the correlation is simply negligible and not of interests. In all cases, we can have a valid condition for this truncation.

% and commonly used family such as Mat\'ern has decreasing $g(\omega)$ over $|\omega|$. 

Using matrix representation of row vector $\mathbf {Q_s}=\{exp(i {\bf s}^T \boldsymbol{\omega}_l)\}_{l=1...m}$ and diagonal matrix ${\bf G} = \{g^{1/2}(\boldsymbol\omega_l)\}_{l,l} $, the construction of (\ref{stationary_construction}) can be viewed as the result of a projection from an $m$-element vector $\bf Y$ in frequency space to an $n$-element location space $\mathbf {Q_s}{\bf G}^{1/2} \bf Y$ and adding random noise. We refer the set $\{\boldsymbol{\omega}_l\}_{l=1...m}$ as the frequency support.

For an finite set of arbitrary locations $\mathbf{S}=\{\mathbf{s}_j\}_{j=1...N}$ with $\mathbf{s}_j\in \mathbb{R}^d$, it can be derived that the joint distribution is:

\begin{equation}
\begin{aligned}
&	\mathbf{Z_S}\sim {\bf N}  (\boldsymbol{0}, {\mathbf {Q_S} }{\mathbf G}\mathbf{Q^*_S}+\mathbf{I}\sigma^2)
\end{aligned}
\label{Eqn:functional_gp}
\end{equation}
where $\bf Q_S$ is the $n$-by-$m$ matrix formed by stacking the row vectors $\bf Q_{S_j}$ and $\mathbf {Q^*_S}$ is its conjugate transpose. It can be derived that the covariance function between any two locations $\mathbf{s}_j,\mathbf{s}_k$ is

\begin{equation}
\begin{aligned}
&	Cov(\mathbf{s}_j,\mathbf{s}_k) = \sum_{l=1}^{m} exp\{i \boldsymbol{\omega}_l^T (\mathbf{s}_j-\mathbf{s}_k)\}g(\boldsymbol{\omega}_l)/n+{\sigma^2}{\bf 1}_{j=k}
\end{aligned}
\label{Eqn:covariance_function}
\end{equation}

We would like to point out that (\ref{Eqn:covariance_function}) forms a valid covariance matrix, as stated by the theorem below. For the conciseness of presentation, all the proofs will be shown in the appendix.

\begin{theorem}
	The covariance matrix formed by (\ref{Eqn:covariance_function}) is real and positive definite.
\end{theorem}

It can also be observed that the function of (\ref{Eqn:covariance_function}) is shift-invariant in $ (\mathbf{s}_j,\mathbf{s}_k)$, therefore $\mathbf{Z_S}$ is weakly stationary. In fact, any weakly stationary Gaussian distribution can be viewed as the limit of formulation in (\ref{Eqn:functional_gp}), due to the following property:

\begin{theorem}
	If we denote a weakly stationary covariance function by $C(\mathbf{s}_j-\mathbf{s}_k)$, and its spectral density as $g(\boldsymbol\omega)=\int_{\mathbb{R}^d} exp(-i  \textbf{x}^T \boldsymbol\omega) C(\textbf{x}) d{\bf x}$. Under mild condition that $g(\boldsymbol\omega)<\epsilon$ if $\boldsymbol\omega \not\in \mathbb{W}$, when  $m= m_1...m_d$  goes to infinity, the function specified by (\ref{Eqn:covariance_function}) converges to the covariance function, that is:
	$$lim_{m\rightarrow \infty} Cov(\mathbf{s}_j,\mathbf{s}_k) =C(\mathbf{s}_j-\mathbf{s}_k)$$
	with the rate of convergence being $O(1/m^2)$.
\end{theorem}

This property is important as under the case of relative large $n_{(.)}$, the spectral construction in (\ref{Eqn:covariance_function}) can be treated as an approximation to a weakly-stationary covariance. However, it would not be fair to view this simply as an approximation method, in fact, when the frequency support set $\{\boldsymbol{\omega}_l\}$ is fixed, the distribution specified in  (\ref{Eqn:covariance_function}) can be extended to uncountable set of locations $\mathbb{R}^d$ and form a valid stochastic process.

\begin{theorem}
	With the frequency support set $\{\boldsymbol{\omega}_l\}_{l=1...m}$ fixed, the finite dimensional distribution of $\mathbf{Z}$ specified by (\ref{Eqn:functional_gp}) satisfies the Kolmogorov consistency criteria:
	\begin{enumerate}
		\item For any finite set $\{\mathbf{s}_1,\mathbf{s}_2,...,\mathbf{s}_n\}$ in $\mathbb{R}^d$ and all of its permutation $\{\mathbf{s}_{\pi1},\mathbf{s}_{\pi2},...,\mathbf{s}_{\pi n}\}$, we have
		$p(\mathbf{s}_{\pi1},\mathbf{s}_{\pi2},...,\mathbf{s}_{\pi n})=p(\mathbf{s}_{1},\mathbf{s}_{2},...,\mathbf{s}_{n})$
		\item For any location in $\mathbf{s}_k \in  \mathbb{R}^d$, we have 
		$p(\mathbf{s}_{1},\mathbf{s}_{2},...,\mathbf{s}_{ n})=
		\int_{ \mathbb{R}^d} p(\mathbf{s}_{1},\mathbf{s}_{2},...,\mathbf{s}_{n},\mathbf{s}_{k}) d\mathbf{s}_{k}$
	\end{enumerate}
\end{theorem}

Therefore, $\mathbf{Z}_{\mathbb{R}^d}$ is a valid stochastic process. We name this process as the functional Gaussian process (FGP).

\subsection{Spectral Dimension Reduction}

We have established the asymptotic equivalence of the functional Gaussian process and the one constructed by the traditional covariance functions. Now we demonstrate its unique advantage in computation.

We first note the properties of the $n$-by-$m$ matrix $\mathbf {Q_S}$ under two conditions:

\begin{theorem}
If the location vector has $\mathbf{S}=\{1,...,n_1\} \times \{1,...,n_2\}\times...\times \{1,...,n_d\}$, and the frequency support $\{\boldsymbol\omega_l\}_l$ has $\Delta_{(.)}=\pi$ and $m(.)\le n(.)$, then we have $ \mathbf{Q^*_S}\mathbf {Q_S} = \mathbf{I}$. When $m(.)=n(.)$,  $ \mathbf{Q_S}\mathbf {Q^*_S} = \mathbf{I}$.
\end{theorem}

This is very important as it greatly reduces the computational burden in the likelihood, with two steps of truncations. We first truncate the frequency support to $(-\pi,\pi)^d$ (that is $\Delta_{(.)}=\pi$ and $m(.)=n(.)$), after ensuring $g(\boldsymbol\omega)<\epsilon$ when $|\omega_{(.)}|>\pi$ in all the directions, for arbitrarily small $\epsilon$. If this condition were not met (although unlikely), we can manually scale up in $\mathbf{S}$ and thereby increase  the range parameter $\rho$, which leads to more rapid decrease of $g(\boldsymbol\omega)$ as discussed in the previous section.

As the second step, we further truncate the rank of $\bf  Q_S$ down to $m(.)<n(.)$, by eliminating the frequencies with $g(\boldsymbol\omega) \ll \sigma^2$. For illustration, we denote the full rank matrix with subscript $n$ and reduced rank with $m$. Since we have $g(\boldsymbol\omega)/\{g(\boldsymbol\omega)+\sigma^2\}={o}(g(\boldsymbol\omega))$ for the omitted $(n-m)$ frequencies, using Woodbury identity  we have $({\mathbf {Q}_n }{\mathbf G_n}\mathbf{Q}_n^*+\mathbf{I}_n\sigma^2)^{-1}=\mathbf{I}_n \sigma^{-2}- \sigma^{-2}{\mathbf {Q}_n } {\bf G}_n({\bf G}_n + \mathbf{I}_n \sigma^{2})^{-1}{\mathbf {Q}_n^* } \approx \mathbf{I}_n \sigma^{-2}- \sigma^{-2}{\mathbf {Q}_m } {\bf G}_m({\bf G}_m + \mathbf{I}_m \sigma^{2})^{-1}{\mathbf {Q}_m^* }$, and $|{\mathbf {Q}_n }{\mathbf G}_n\mathbf{Q}_n^*+\mathbf{I}\sigma^2|= |{\bf G}_n+\mathbf{I}_n\sigma^2|\approx|{\bf G}_m+\mathbf{I}_m\sigma^2||\mathbf{I}_{(n-m)}\sigma^2|$. Our empirical finding is that truncation at $g(\boldsymbol{\omega}) \ge 0.01 \cdot \sigma^2$ results in indistinguishable parameter estimation. More will be discussed in the simulation studies.

\subsection{Bayesian Modeling and Posterior Computation}

Similar to other spectral methods (e.g. \cite{stroud2014bayesian}), we utilize lattice embedding to obtain the frequency estimates. Before we elaborate the method, it is worth pointing out that the lattice latent variable is only an auxiliary tool for posterior estimation, provided a finite set of data are collected. This does not contradict the definition of functional Gaussian process on any continuous domain. In fact, unlike other lattice method, we can have even multiple observations on the same location and our construction is still valid.

We now assume the data $\mathbf {\tilde{Z}_{\tilde S}}$ are originally observed on $\tilde{n}$ coordinates $\tilde {\mathbf S}$. We also assume there is a latent random variable $\mathbf {Z_S}$ on the lattice $\mathbf{S}=\{1,...,n_1\} \times \{1,...,n_2\}\times...\times \{1,...,n_d\}$. To satisfy the two conditions, we divide $\tilde {\mathbf S}$ by the greatest common divisor in each direction of the $\{\tilde{\mathbf s}_j- \tilde{\mathbf s}_k\}$ and shift the smallest coordinate to $(1,1,...,1)$. We denote the transformed set as $\mathbf{S_o}$, so we have $\mathbf{S_o}\subset \mathbf{S}$. As it is common to relax the homogeneity assumption about the random noise, we added a diagonal matrix with location specific random noise  $\mathbf{V_{S_o}}$. We have:

\begin{equation}
\begin{aligned}
&	\mathbf{\tilde{Z}_{S_o}}|\mathbf{Z_S}\sim {\bf N}(\mathbf{Z_{S_o}},\mathbf{V_{S_o}})  \\
&	\mathbf{Z_S}\sim {\bf N}  (\boldsymbol{0}, {\bf Q_S} {\bf G}\mathbf{Q_S}^*+\mathbf{I}\sigma^2)
\end{aligned}
\label{Eqn:lattice_prior}
\end{equation}
where $\mathbf{V_{S_o}}=\{\nu_{i}\}$ is a positive diagonal matrix.

The augmented likelihood becomes a product of $m+\tilde{n}$ independent normal density:

\begin{equation}
\begin{aligned}
p(\mathbf{\tilde{Z}_{S_o}},\mathbf{Z_S})
&\propto  \prod_{l=1}^{m} exp \Big(		-\frac{1}{2} [  \frac{{{\bf (Q^*Z)}_l { (\bf QZ)}_l}}{\{g(\boldsymbol{\omega}_l)+\sigma^2\}} + log\{g(\boldsymbol{\omega}_l)+\sigma^2\}]\Big)\\
\cdot&\prod_{i=1}^{\tilde{n}} exp \Big(-\frac{1}{2} \{ \frac {(\tilde{Z}_{S_oi}-Z_{S_oi})^2}{\nu^2_{i}}+log(\nu^2_{i})\}\Big)
\\
\end{aligned}
\label{Eqn:fgp_likelihood}
\end{equation}
where ${\bf (Q^*Z)}_l$ denotes the $l$th element $\bf Q^*Z$. We would like to point out the famous Whittle's likelihood \citep{whittle1953analysis} is a special case of the first step truncation to $(-\pi,\pi)^d$ in dimension $n$ on a full lattice. Here we present a more general case with further dimension reduction from $n$ to $m$. The product of $\mathbf{Q^*}\bf Z$ corresponds to the $m$-element truncated inverse Fourier transform of $\bf Z$. The full computation can be carried out at a complexity of $O(n \log_2 n)$ using Fast Fourier Transform (\cite{cooley1965algorithm}) or even faster at $O(m \log_2 n)$ with the recently invented Sparse Fourier Transform \citep{hassanieh2012nearly}.

For posterior sampling, we use Gibbs sampling from the individual full conditional distribution. However, it is computationally demanding to generate random samples from $\bf Z_S|\tilde{Z}_{S_o}$. In a similar study of stationary Gaussian process, \cite{stroud2014bayesian} proposed using a $k$-iteration solver at each step with a complexity at $O(k n \log_2 n)$. Here we introduce a more efficient sampling scheme at $O( n \log_2 n)$ with the latent lattice variable $\boldsymbol\mu_{\bf} = \bf Q_S G^{1/2} Y$ followed by $\bf{Z_S}\sim {\bf N}  (\boldsymbol{\mu}, {\bf I}\sigma^2)$.

We denote the covariance parameter in $\bf G$ as $\boldsymbol\theta$ and its prior distribution as $p(\boldsymbol{\theta})$. As the covariance function has the general form of $C(x|\boldsymbol\theta)=\theta_1 h(x/\theta_2)$, to assure posterior propriety, we assign proper diffuse prior, $IG(0.1,0.1)$ for the scale parameter $\theta_1$ and uniform prior $U(0, 1000)$ for the covariance parameter $\theta_2$. If a stricter condition can be met such that $\nu_i=\nu_j$ for any $i,j$, the objective improper prior such as \cite{berger2001objective} is more desirable. The full conditional distributions can be derived:

\begin{equation}
\begin{aligned}
\bf Y| {\bf Z_S} &\sim  CN ({\bf G^{1/2}(G+I\sigma^2) ^{-1}Q_S^*Z_S, \sigma^2(G+I\sigma^2)^{-1}}),\\
\boldsymbol\mu_{\bf} = &\bf Q_S G^{1/2} Y,\\
p(\boldsymbol{\theta}|{\bf Z}) &\propto p({\bf Z|\boldsymbol{\theta}})p(\boldsymbol{\theta}),\\
{Z}_i| {\mu}_i,&\tilde{ Z}_i \stackrel{indep}{\sim}N((\frac{1}{\sigma^2}+\frac{1}{\nu_i^2})^{-1}(\frac{\mu_i}{\sigma^2}+\frac{\tilde{Z}_i}{\nu_i^2}), (\frac{1}{\sigma^2}+\frac{1}{\nu_i^2})^{-1}) \text{, if $i\in \bf S_o$},\\
{Z}_i|{\mu}_i & \stackrel{indep}{\sim}  { N}  ({\mu_i}, \sigma^2 ) \text{, if $i \notin \bf S_o$}.
\end{aligned}
\label{fgp_conditional}
\end{equation}

\subsection{Predictive Distribution}

It is a common use of Gaussian process to make prediction at a set of arbitrary locations, which is known as Kriging \citep{cressie1988spatial}. The functional Gaussian process also accommodates this demand, since the covariance matrix between two sets $\bf S_1$ and $\bf S_2$ is simply ${\bf Q_{S_1}} {\bf G}\mathbf{Q^*_{S_2}}$. The predictive distribution is
$	\mathbf{\tilde{Z}_{S_2}|\tilde{Z}_{S_1}}\sim {\bf N}   ( {\bf Q_{S_2}} {\bf G}\mathbf{Q_{S_1}}^*({\mathbf {Q_{S_1}} }{\mathbf G}\mathbf{Q^*_{S_1}}+\bf I\sigma^2+\mathbf{V_{S_1}})^{-1}{\bf \tilde{Z}_{S_1}}, 
 {\bf Q_{S_2}} {\bf G}\mathbf{Q_{S_2}}^* +\bf I\sigma^2+ \mathbf{V_{S_2}}-
{\bf Q_{S_2}} {\bf G}\mathbf{Q_{S_1}}^*({\mathbf {Q_{S_1}} }{\mathbf G}\mathbf{Q^*_{S_1}}+\bf I\sigma^2+\mathbf{V_{S_1}})^{-1}
{\bf Q_{S_1}} {\bf G}\mathbf{Q_{S_2}}^*)$. To facilitate the computation, when conditioning on the latent variable $\bf Y$ in the frequency space, the predictive distribution can be further simplified to:

\begin{equation}
\begin{aligned}
\mathbf{\tilde{Z}_{S_2}|Y}\sim {\bf N}  ( {\bf Q_{S_2}} {\bf G^{1/2} Y}, \bf V +I\sigma^2),
\end{aligned}
\label{predictive_distribution_1}
\end{equation}
which is very efficient due to the independent condition.

%Last but not least, it is worth pointing out there is great sparsity in the spectral density $\bf G$, which has been a major driving factor for study in the frequency space. Its estimation is therefore of great interests. For this purpose, the functional Gaussian process provides two estimates: the parametric one via the function of $g$ and the nonparametric one via the the Euclidean norm of each element in ${\bf Q^*_{S}{\bf Y_{S}}}$, which is also known as periodogram. We will illustrate the use of these estimates in the data application.

\subsection{Misaligned Model for Lower Resolution Analysis}

It is worth noting that, the dimension reduction in our method is a result of the inherent sparsity in the frequencies, as opposed to the approximation at the lower resolution. In fact, all the latent variables in the location space are modeled at the same resolution as the data themselves. Therefore, FGP provides estimation directly on the finest resolution. Thanks to the high computational efficiency, it is easy to set up latent variables on a large lattice without costing much, since the time demand only increases linearly. 

On the other hand, there are scenarios where one may be interested in a lower resolution analysis. For examples, some observations might be very close and there is little gain to measure their correlation (since it is close to $1$); some coordinates may contain measurement errors and therefore perfectly assigning them to a dense lattice is unnecessary; there might be a need for a even more real-time computation.

In these cases, we propose the approximation model for FGP. Assume a collection of random variables from a functional Gaussian process are located on a lattice $\bf S$, we collect data at ${\mathbf x}_i$ near the lattice point. To assign one lattice point for one ${\mathbf x}_i$, we define ${\bf s}^*_i=min_{\bf s_j}argmin_{\bf s_j\in \bf S}||\bf x_i-s_j||$. As a result, each lattice point ${\bf s_j}$ may have multiple data assigned, we denote the set as $\chi_{\bf {\bf s_j}}=\{\text{all }{\bf x}_i \text{ s.t. } {\bf s}^*_i={\bf s}_j\}$

To model these data, we simply treat them as misaligned: we assign the value on the closest lattice point ${\mathbf s}_j$ as the mean and an increasing function in the distance $||{\mathbf x}_i-{\mathbf s}_j||$ as the variance. That is,

\begin{equation}
\begin{aligned}
&	\mathbf{Z_S}\sim {\bf N}  (\boldsymbol{0}, {\bf Q_S} {\bf G}\mathbf{Q^*_S}+\mathbf{I}\sigma^2),\\
&	{\tilde{Z}_{\bf x_i}}|{Z_{\bf s_j}}\stackrel{indep}{\sim} { N}({Z_{\bf s_j}},\lambda({||\bf x_i-\bf s_j||})) \text { for all ${\bf x_i\in\chi_{\bf {\bf s_j}}}$}.
\end{aligned}
\label{Eqn:misaligned_model}
\end{equation}
One example of such function is $\lambda({||\bf x_i-\bf s_j||}))=\nu^2_i(1+\kappa {||\bf x_i-\bf s_j||})$, for which if ${||\bf x_i-\bf s_j||}=0$,  this reduces to the formulation in (\ref{Eqn:lattice_prior}). Therefore, the misaligned model is a generalized case of (\ref{Eqn:lattice_prior}), the properties and sampling algorithm of functional Gaussian process stills apply in this scenario. The only modification is the posterior  ${Z_{\bf s_j}}| {\mu}_{{\bf s_j}},\{\tilde{Z}_{\bf x_i}\}_{x_i\in\chi_{\bf {\bf s_j}}} \stackrel{indep}{\sim}N((\frac{1}{\sigma^2}+\sum\frac{1}{\nu_i^2})^{-1}(\frac{{\mu}_{{\bf s_j}}}{\sigma^2}+\sum\frac{\tilde{Z}_{\bf x_i}}{\nu_i^2}), (\frac{1}{\sigma^2}+\sum\frac{1}{\nu_i^2})^{-1})$.

\section{Non-stationary Functional Gaussian Process}

We now propose a more general framework that gives arise to a non-stationary Gaussian process. Our work is largely inspired by the pioneer work in evolutionary spectrum and semi or  locally stationary process (\cite{priestley1965evolutionary}). The former is to let the spectral process change with location $H_{\bf s} =  A({\bf s},{\boldsymbol \omega})  Y({\boldsymbol \omega})$, where $Y({\boldsymbol \omega})\sim CN(0, 1) \text{  for each $\boldsymbol{\omega}$}$. As the modulating function $A({\bf s,{\boldsymbol \omega})}$ changes with location $\bf s$, the process and its Fourier transform  $ \tilde{Z}_{\bf s}=\int_{\omega\in\mathbb{R}^d} exp(i {\bf s}{\boldsymbol \omega}) H_{\bf s}({\boldsymbol \omega})d {\boldsymbol \omega}$ become non-stationary. The latter is defined in the sense that $A({\bf s},{\boldsymbol \omega})$ is slowly varying with ${\bf s}$ such that within a small region, the process can be treated as stationary. We now combine this notion with the recently popularized mixture modeling framework to construct a new non-stationary process.

%This theoretic gap motivates us to develop a new solution with the inter-region dependency and the probabilistic justification. 

\subsection{A Stick-Breaking Process with Spectral Dependency}

%$$& \tilde{Z}_{\bf s}=\int_{\omega\in\mathbb{R}^d} exp\{i {\bf s}{\boldsymbol \omega}\} H_{\bf s}({\boldsymbol \omega})d {\boldsymbol \omega} \\$$

We first construct a non-stationary spectral process $H_{\bf s}({\boldsymbol \omega})$ on frequency $\boldsymbol{\omega}\in\mathbb{R}^d$. At location $\bf s$:
\begin{equation}
\begin{aligned}
& H_{\bf s}({\boldsymbol \omega}) =  T_k({\boldsymbol \omega})\text{ with probability $p_{k,{\bf s}}$},\\
& T_k({\boldsymbol \omega})= g_k^{1/2}(\boldsymbol\omega)Y(\boldsymbol{\omega}),\\
& p_{k,{\bf s}}=u_{k,{\bf s_i}}\prod_{j<k}(1-u_{j,{\bf s}}),
\label{nonstationary_spectral_process}
\end{aligned}
\end{equation}
where $g_k(\boldsymbol\omega)$ is the spectral density function of the $k$th certain class; $ Y(\boldsymbol\omega)\sim CN(\bf 0,I)$ is a complex normal vector folded over $0$, as defined before; $p_{k,{\bf s}}$ is the stick-breaking weight that varies in $\bf s$. Similar to \cite{priestley1965evolutionary}, the non-stationarity is realized via on different modulating functions $A({\bf s,{\boldsymbol \omega})}= g_k^{1/2}({\boldsymbol \omega})$; the distinction is that this function is now regulated via a stick-breaking process.

The uniqueness of this stick-breaking process is that there is only one copy of $Y(\boldsymbol\omega)$, shared by finite or infinite many components. This leads to dependent complex normal distribution for $T_{k_1}({\boldsymbol \omega}),T_{k_2}({\boldsymbol \omega}) \sim CN(0, g_{k_1}^{1/2}g_{k_2}^{1/2})$ for $\boldsymbol{\omega}>0$, even if $k_1\ne k_2$.

\subsection{Non-stationary Functional Gaussian Process}

Similar to stationary functional Gaussian process, given the finite fixed frequency support $\boldsymbol{\omega}\in \mathbb{W}$, we define the process in the continuous data domain:

\begin{equation}
\begin{aligned}
{Z}_{\bf s}=  {\bf Q}_{\bf s} H_{\bf s}+\xi_{\bf s} \text{,   where $\xi_{\bf s}\sim N(0,\sigma^2)$},
\label{ns_fgp}
\end{aligned}
\end{equation}
where $H_{\bf s}$ is defined in (\ref{nonstationary_spectral_process}). The distributional differences of $H_{\bf s}$ is controlled by the location varying $p_{k,{\bf s}}$.  Marginalized over $p$, the covariance between two locations is $Cov({Z_{\bf s_1}}, Z_{\bf s_2})={\mathbf {Q_{\bf s_1}} } (\sum_{k_1=1}^{\infty} \sum_{k_2=1}^{\infty} p_{{k_1,{\bf s_1}}}p_{k_2,{\bf s_2}}\mathbf G^{1/2}_{k_1}\mathbf G^{1/2}_{k_2})  \mathbf{Q^*_{\bf s_2}}+ \sigma^2 1_{{\bf s}_1={\bf s}_2}$.
More importantly, conditional on the latent class assignment $C_{\bf s}$, all the observations on $\bf S$ are correlated and jointly form a multivariate Gaussian distribution with mean $\bf 0$ and covariance:

\begin{equation}
\begin{aligned}
& Cov ({Z_{\bf s_1}}, Z_{\bf s_2}| C_{\bf s_1},C_{\bf s_2})= {\mathbf {Q_{\bf s_1}} } (\mathbf G^{1/2}_{C_{\bf s_1}}\mathbf G^{1/2}_{C_{\bf s_2}})  \mathbf{Q^*_{\bf s_k}}+ \sigma^2 1_{{\bf s}_1={\bf s}_2}.
\end{aligned}
\label{nonstationary_covariance}
\end{equation}

We again verify the requirements stated in the following theorem.

\begin{theorem}
For finite set ${\bf S}=\{s_1,s_2,...,s_n\}$, The covariance function in (\ref{nonstationary_covariance}) generates a positive-definite matrix. The finite dimensional density of $\bf Z_{\bf S}$ satisfies the Kolmogorov consistency criteria and therefore can extend to a valid stochastic process $\bf Z_{\mathbb{R}^d}$ .
\end{theorem}

Therefore, we refer (\ref{ns_fgp}) as non-stationary functional Gaussian process (NS-FGP). The conditional dependency in NS-FGP is inherited from the shared spectral vector $Y$, despite of the different projections into the location space. This is a major distinguishing factor of our method from the other stick-breaking constructions like \cite{duan2007generalized}, where $Cov ({Z_{\bf s_1}}, Z_{\bf s_2}| C_{\bf s_1},C_{\bf s_2})=0$ if $C_{\bf s_1} \ne C_{\bf s_2}.$

There are different choices to induce location varying $p$, such as hidden Markov random field \citep{franccois2006bayesian}, generalized spatial Dirichlet process \citep{duan2007generalized}, probit transform of a Gaussian process \citep{rodriguez2011nonparametric}. For our purpose, the last model is especially appealing for two reasons: first, the moderately large magnitude of $\mu$ (e.g. $|\mu|>3|$) in the probit link $p=\phi(\mu)$ can generate $p$ close to 0 or 1, hence much less randomness in $C$ and a clearer clustering pattern; second, this framework can be easily extended with our stationary functional Gaussian process to have extremely fast sampling speed. We describe the model for $p$ as follows:

\begin{equation}
\begin{aligned}
& p_{k,{\bf s}}=u_{k,{\bf s_i}}\prod_{j<k}(1-u_{j,{\bf s}}) ,\\
& u_{k,{\bf s}} = Pr (L_{k,{\bf s}}\ge 0) ,\\
& {\bf L}_{k,{\bf S}} \sim {\bf N}  ({\bf 0}, {\mathbf {Q_{\bf S}} }{\mathbf M}_{k}\mathbf{Q^*_{\bf S}}+{\bf I}),\\
\end{aligned}
\label{weight_process}
\end{equation}
where ${\bf L}_{k,{\bf S}}$ is assumed to be from an FGP and ${\mathbf M_k}$ is a diagonal matrix formed by a certain spectral density function.

\subsection{Posterior Computation}

We use the following data augmentation scheme to facilitate the posterior sampling for NS-FGP. To allow for more flexible consideration, we again relax the assumption about the homogeneous random error:
\begin{equation}
\begin{aligned}
&{\tilde {Z}}_{\bf s} |\{C_{s}=k\} = Z_{{\bf s},k} + \xi_{{\bf s}} \text{ ,where $\xi_{{\bf s}} \sim N(0,\nu_s^2)$},\\
& C_{\bf s}=k  \text{ w.p. $p_{{\bf s},k}$},\\
&\mathbf{Z}_{{\bf S},k}\sim {\bf N}  (\boldsymbol{\mu}_{{\bf S},k}, \mathbf{I}\sigma_k^2),\\
&\boldsymbol{\mu}_{{\bf S},k}={\bf Q}_{\bf S} {\bf G}^{1/2}_k{\bf Y} ,\\
& u_{k,{\bf s}} = p (L_{k,{\bf s}}\ge 0) ,\\
& {\bf L}_{k,{\bf S}} \sim {\bf N}  ({\bf 0}, {\mathbf {Q_{\bf S}} }{\mathbf M}_{k}\mathbf{Q^*_{\bf S}}+{\bf I}).\\
\end{aligned}
\end{equation}

Then the augmented likelihood-prior probability is:

\begin{equation}
\begin{aligned}
Normal({\bf \tilde{Z}_{S_o}|Z_S, C_S}) \times FGP({\bf Z_S| C_S}) \times SB({\bf C_S}|{\bf L_{k,S}})\times FGP({\bf L_{k,S}})\times Prior(\boldsymbol\theta_G,\boldsymbol\theta_M, \sigma^2)
\end{aligned}
\end{equation}
where $Normal$ is the density for independent normal distribution, $FGP$ is the one for stationary functional Gaussian process, $SB$ is the one for stick-breaking process. Similar to the stationary case, we assign all the covariance parameters as $U(0,1000)$ and all the scale parameters as $IG(0.1,0.1)$. The full conditional distributions of the two FGPs are mutually independent with computational complexity $O(m \log_2 n)$. We list the sampling algorithm as follows:

\begin{enumerate} 
	\item Draw $C_{\bf s_i}$ from {$\{1,2,...\}$} from\\ $p(C_{\bf s_i}=k) \propto {\bf 1}({w_{k,\bf s_i}>r_{\bf s_i}}) Normal(Z_{\bf s_i} |Y_{k,{\bf s_i}}, \nu_{\bf s_i})$.
	\item Sample from $p(r_{\bf s_i}) = Uniform(0, p_{C_{\bf s_i},\bf s_i})$	.
	\item Sample from $p({L}_{k,{\bf s_i}})= Normal({\eta}_{k,{\bf s_i}},1){\bf 1}({\eta}_{k,{\bf s_i}}\ge 0, C_{\bf s_i}=k) +Normal({\eta}_{k,{\bf s_i}},1){\bf 1}({\eta}_{k,{\bf s_i}}<0, C_{\bf s_i} > k)$. And $p({L}_{k,{\bf s_i}})= Normal({\eta}_{k,{\bf s_i}},1)$ if $C_{\bf s_i}<k$ or $C_{\bf s_i}$ is unobserved.
	\item Sample from $p({\boldsymbol\theta_{M,k}} )\propto FGP ({L}_{k,{\bf s_i}}|\boldsymbol{\theta}_{M,k})\pi(\boldsymbol{\theta}_{M,k})$.
	\item Sample from $p(\boldsymbol \eta_{k, \bf S}) = MVN({{\bf Q_SM}_k({\bf M}_k+{\bf I}) ^{-1}{\bf Q_S^*}{\bf L_{k,S}}, {\bf Q_S}({\bf M}_k-{\bf M}_k({\bf M}_k+{\bf I})^{-1}{\bf M}_k){\bf Q_S^*}})$.
	\item Compute $u_{k,\bf s_i}=\Phi(\eta_{k,\bf s_i})$ and $w_{k,\bf s_i}= u_{k,\bf s_i}\prod_{k<l}(1-{u_{k,\bf s_i}} )$.
		\item Sample from $p({\boldsymbol\theta_{Z,k}}) \propto FGP ({\bf Z_{\{S:C_{s}=k\}}}|\boldsymbol{\theta}_{Z,k})\pi(\boldsymbol{\theta}_{Z,k})$.
		\item Sample from
		$p(\boldsymbol Y)= CN(({\bf I} + \sum_k {\bf G}_k \sigma_k^{-2})^{-1}( \sum_k\sigma_k^{-2}{\bf G}_k^{1/2} {\bf Q}^* {\bf Y}_k), ({\bf I} +\sum_k {\bf G}_k \sigma_k^{-2})^{-1})$.
		\item Compute $\boldsymbol{\mu}_{{\bf S},k}={\bf Q}_{\bf S} {\bf G}^{1/2}_k{\bf Y}$ .
		\item Sample from
		 $p({Z}_{s,k})=Normal((\frac{1}{\sigma_k^2}+\frac{1}{\nu_i^2})^{-1}(\frac{\mu_i}{\sigma_k^2}+\frac{\tilde{Z}_i}{\nu_i^2}), (\frac{1}{\sigma_k^2}+\frac{1}{\nu_i^2})^{-1})$
		 if $C_s=k$; else
		 $p({Y}_{s,k})=Normal  ({\mu_{s,k}}, \sigma_k^2 )$.
\end{enumerate}

Similar to stationary FGP, the predictive distribution for NS-FGP is $	\mathbf{\tilde{Z}_{S_2}|\tilde{Z}_{S_1}, G_{S_2}}\sim {\bf N}   ( {\bf Q_{S_2}} {\bf G_{S_2}G_{S_1}}\mathbf{Q_{S_1}}^*({\mathbf {Q_{S_1}} }{\mathbf G}\mathbf{Q^*_{S_1}}+\bf I\sigma^2+\mathbf{V_{S_1}})^{-1}{\bf \tilde{Z}_{S_1}}, 
{\bf Q_{S_2}} {\bf G}\mathbf{Q_{S_2}}^* +\bf I\sigma^2+ \mathbf{V_{S_2}}-
{\bf Q_{S_2}} {\bf G}\mathbf{Q_{S_1}}^*({\mathbf {Q_{S_1}} }{\mathbf G}\mathbf{Q^*_{S_1}}+\bf I\sigma^2+\mathbf{V_{S_1}})^{-1}
{\bf Q_{S_1}} {\bf G}\mathbf{Q_{S_2}}^*)$, or simply conditional on the spectral vector $\mathbf{\tilde{Z}_{S_2}|Y,G_{S_2}}\sim {\bf N}  ( {\bf Q_{S_2}} {\bf G_{S_2}^{1/2} Y}, \bf V +I\sigma^2)$,where the latent $\bf G_{S_2} =\bf G_{k}$ with probability $p_{S_2,k}$.

\section{Data Applications}

We now demonstrate the use of functional Gaussian process via simulated and real data applications.

\subsection{Simulated Data}

We first assess the performance of parameter estimation in stationary FGP. We generated 2,500 locations randomly inside a square space ${\bf s}_i=(x_{i1},x_{i2})\sim U(0,100) \times U(0,100)$ (Figure\ref{sim1}(a)). As most of the existing spatial packages assume isotropic covariance, for comparison, we first tested the two isotropic functions:(1)  Mat\'ern function with the smooth degree at $\kappa=1.5$,  $C({\bf s}_i-{\bf s}_j)= \phi (1+{||{\bf s}_i-{\bf s}_j||/{\rho_1}}) exp(-{||{\bf s}_i-{\bf s}_j||/{\rho_1}})$; (2) squared exponential function or sometimes called ``Gaussian'' covariance, $C({\bf s}_i-{\bf s}_j)= \phi exp(-{||{\bf s}_i-{\bf s}_j||^2/{2\rho_1^2}})$, which can be viewed as Mat\'ern function with the smooth degree at $\kappa=\infty$  \citep{stein1999interpolation}. 

We would like to point out that the degree of smoothness $\kappa$ is associated with the degree of  sparsity in the spectral density matrix $\bf G$ in FGP. In $d$-dimension, the spectral density of Mat\'ern family is $g(\boldsymbol\omega)\propto (\rho^2+||\boldsymbol\omega||_d^2)^{-\kappa-d/2}$. Therefore, the spectral density with larger value of $\kappa$ will approach 0 faster. As illustrated in Figure\ref{sim1}(b), the squared exponential is much closer to 0 (hence much sparser) than Mat\'ern $\kappa=1.5$ as $|\omega|$ increases. Using our empirical truncation criterion $g(\boldsymbol{\omega}) \ge 0.01 \cdot \sigma^2$, we found that in Mat\'ern $\kappa=1.5$,  $\rho_1 = 5$ did not lead to any truncation at all; whereas in squared exponential, the spectral density is truncated at $m=13\%n$, which leads to almost 8 times complexity reduction.

We compare the results against various spatial methods available in R, such as the maximum likelihood method (``geoR''), traditional Bayesian Gaussian process (``spBayes'') and predictive process (``spBayes''). The results are listed in Table~\ref{statioanry_comparison}.

For Mat\'ern $\kappa=1.5$ , FGP provides the most accurate estimate for the parameters and is the fastest method among all, thanks to its spectral algorithm. Gaussian predictive process (GPP) \citep{banerjee2008gaussian} is also very efficient in computation due to its dimension reduction, but its estimate for the range parameter $\rho_1$ deviates from the true value. This result is consistent with the recent work of \cite{datta2014hierarchical}, whose nearest-neighbor Gaussian process method showed great improvement over predictive process in parameter estimation.

For squared exponential function, we see a computational time reduction in GPP, due to the lower evaluation cost of the covariance function. On the other hand, the sparsity in the spectral density suggests that the covariance function is ill-conditioned for matrix inversion. This severely affected parameter estimates in other methods: especially in the last two Bayesian methods, where we had to use smaller upper bound in the uniform prior of $\rho$ to avoid the the matrix singularity. Nevertheless, FGP is not susceptible to this issue, since no matrix inversion is involved. The sparse values in $g$ enables us to test the reduced dimension FGP. As shown in the Table~\ref{statioanry_comparison}, the results show almost no difference, while the computation time is reduced by 6 times.

\begin{table}[ht]
	\centering
	\tiny
	\begin{tabular}{ l l |l | l | l | l | l }
		\hline
		& &	FGP (m=n) & 	FGP $(m \ll n)$& MLE & Full Bayes GP & GPP (64 knots)  \\
		\hline
			
								\hline
				
		Mat\'ern with $\kappa=1.5$  & $\rho_1 = 5$ & $5.01 (0.10)$ &  & $ 4.78 (0.23)$ &  $ 4.85 (0.50)$ & $7.46 (0.47)$  \\
		& $\phi = 100$ & $98.00 (16.50)$   && $86.56 (10.54)$& $ 82.33 (26.67)$& $127.95 (6.71)$ \\
		& $\sigma^2 = 1$ & $0.60 (0.21)$ & & $1.1 (0.51)$ & $ 0.93 ( 0.10)$ & $1.16 (0.84)$ \\
		& Time & 277 secs && 547 secs & 31359 secs&  605 secs \\
				\hline

										Squared Exponential & $\rho_1 = 5$ & $ 4.83 (0.08)$  &  $ 4.87 (0.17)$ & $ 6.98 (0.23)$ &  $ 2.87(0.49)$ & $ 2.66 (0.50)$  \\
										& $\phi = 100$ & $ 98.79 (8.80)$  & $ 95.79 (6.80)$& $100.70 (14.57)$& $ 103.25(6.68)$& $105.95 (7.71)$ \\
										& $\sigma^2 = 1$ & $0.50 (0.13)$ & $ 0.53 (0.12)$ & $1.00 (0.44)$ & $ 0.96 ( 0.11)$ & $1.52 (0.28)$ \\
										& Time & 237 secs & 40 secs & 759 secs & 2103 secs&  257 secs \\
															\hline
	\end{tabular}
	\caption{Comparison of estimation for isotropic Gaussian process}
	\label{statioanry_comparison}
\end{table}

\begin{figure}[!ht]
	\centering
		\begin{subfigure}[t]{.45\columnwidth}
		\centering\includegraphics[width=1\columnwidth]{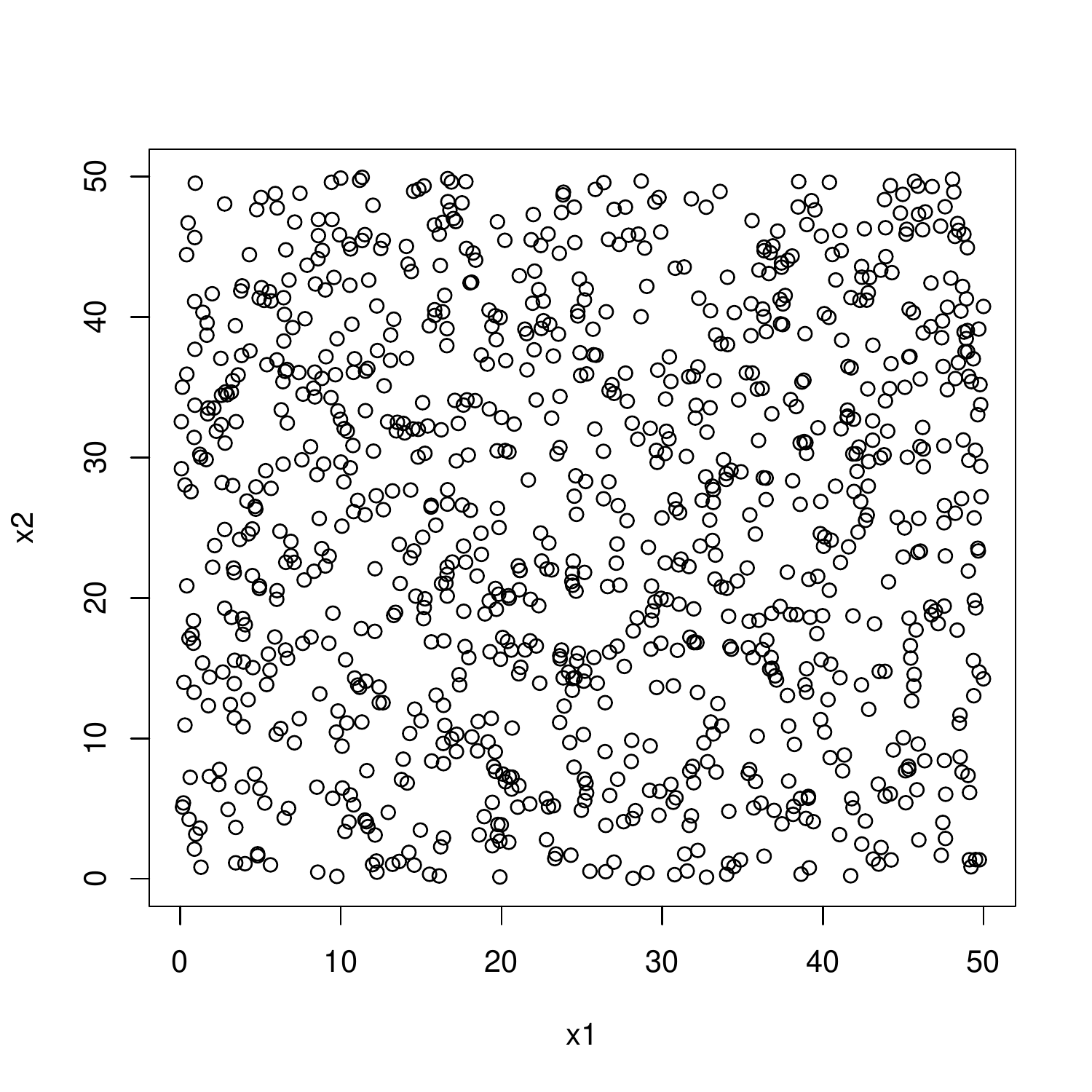}
		\subcaption{Locations of the simulated data}
				\end{subfigure}
		\begin{subfigure}[t]{.45\columnwidth}
	\centering\includegraphics[width=1\columnwidth]{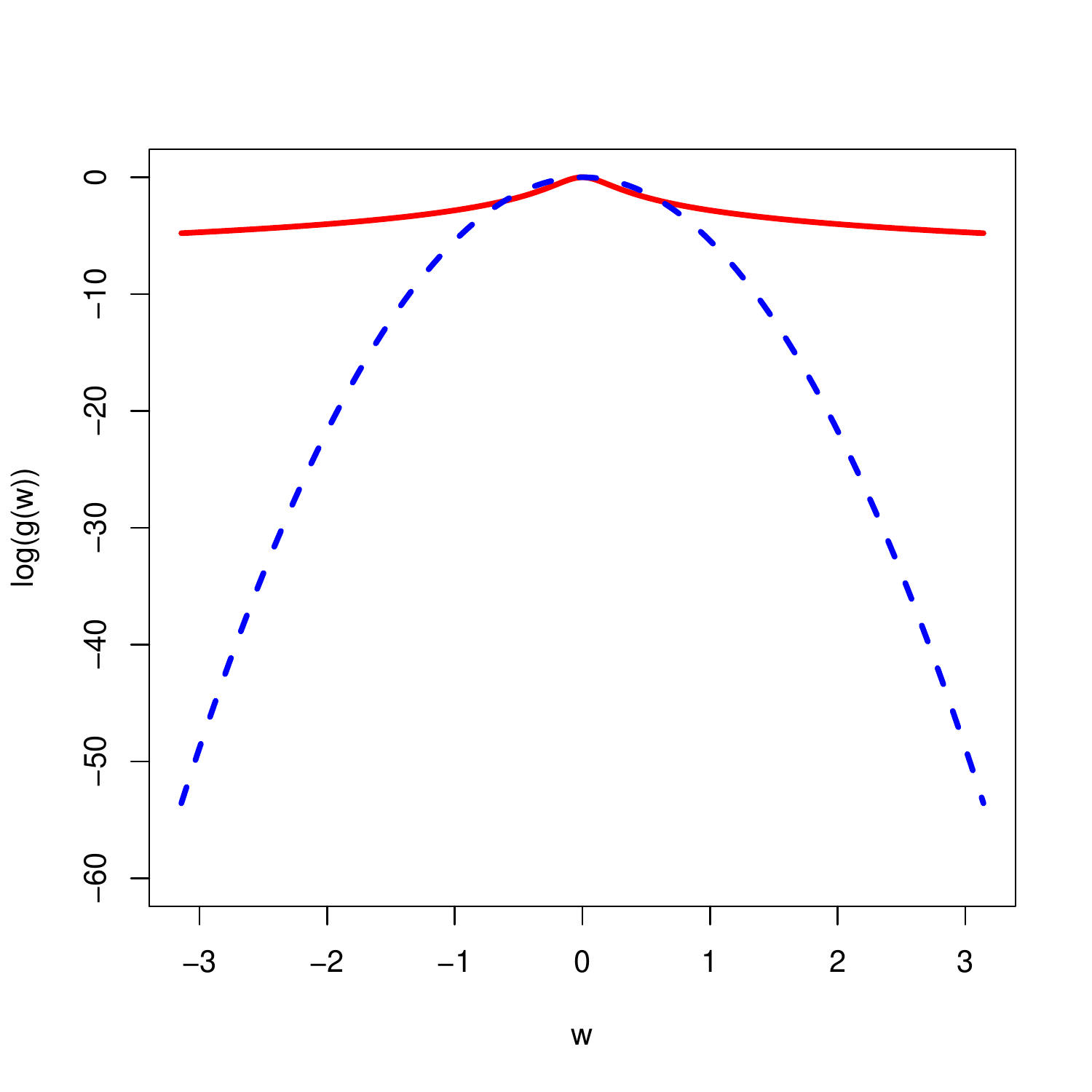}
			\subcaption{Two spectral densities at $\log_{10}$ scale: squared exponential (dashed) decreases  much faster than Mat\'ern (solid).}
		\end{subfigure}
		\caption{Simulation for Stationary FGP}
		\label{sim1}
\end{figure}

We now assess the performance of the non-stationary FGP. To simulate non-stationary surface data, we use the the formulation in \cite{pintore2004spatially}. We use a localized squared exponential covariance with $C({\bf s_1, s_2})= \phi h_{\bf s_1, s_2}exp(-||{\bf s_1- s_2}||^2/\alpha_{\bf s_1, s_2})$, where $\alpha_{\bf s_1, s_2}=(\alpha({\bf s_1})+\alpha({\bf s_2)})/2$ and $h_{\bf s_1, s_2} =2\alpha({\bf s_1})^{1/2}\alpha({\bf s_2})^{1/2}/(\alpha({\bf s_1})+\alpha({\bf s_2}))$ and $\alpha({\bf s_i})=2\rho(\bf s_i)^2$. To assign values to the local range parameter, we use a smooth surface from the function $\rho(s_1, s_2)= (cos(4\pi s_1/100)+2) exp(s_2/200)$ with $s_1,s_2 \in (0,100)^2$ (Figure~\ref{sim2}(a)), which generates data with different range of correlation (Figure~\ref{sim2}(e)). The NS-FGP model converges to 3 dominating clusters, with distribution pattern highly resembles the one for the range parameter (Figure~\ref{sim2}(b,c,d)). The estimated range parameters  correspond to different correlation strength (Figure~\ref{sim2}(f,g,h)). To test the prediction performance, we reran the model with a random $80\%$ sample of the data and predict on the remaining set. The non-stationary FGP model outperforms both the stationary FGP and the GPP models in cross validation metrics of root-mean-square error (RMSE) and median absolute deviation (MAD).

\begin{figure}[!ht]

	\centering
	
		\begin{subfigure}[t]{.24\columnwidth}
			\centering\includegraphics[width=1\columnwidth]{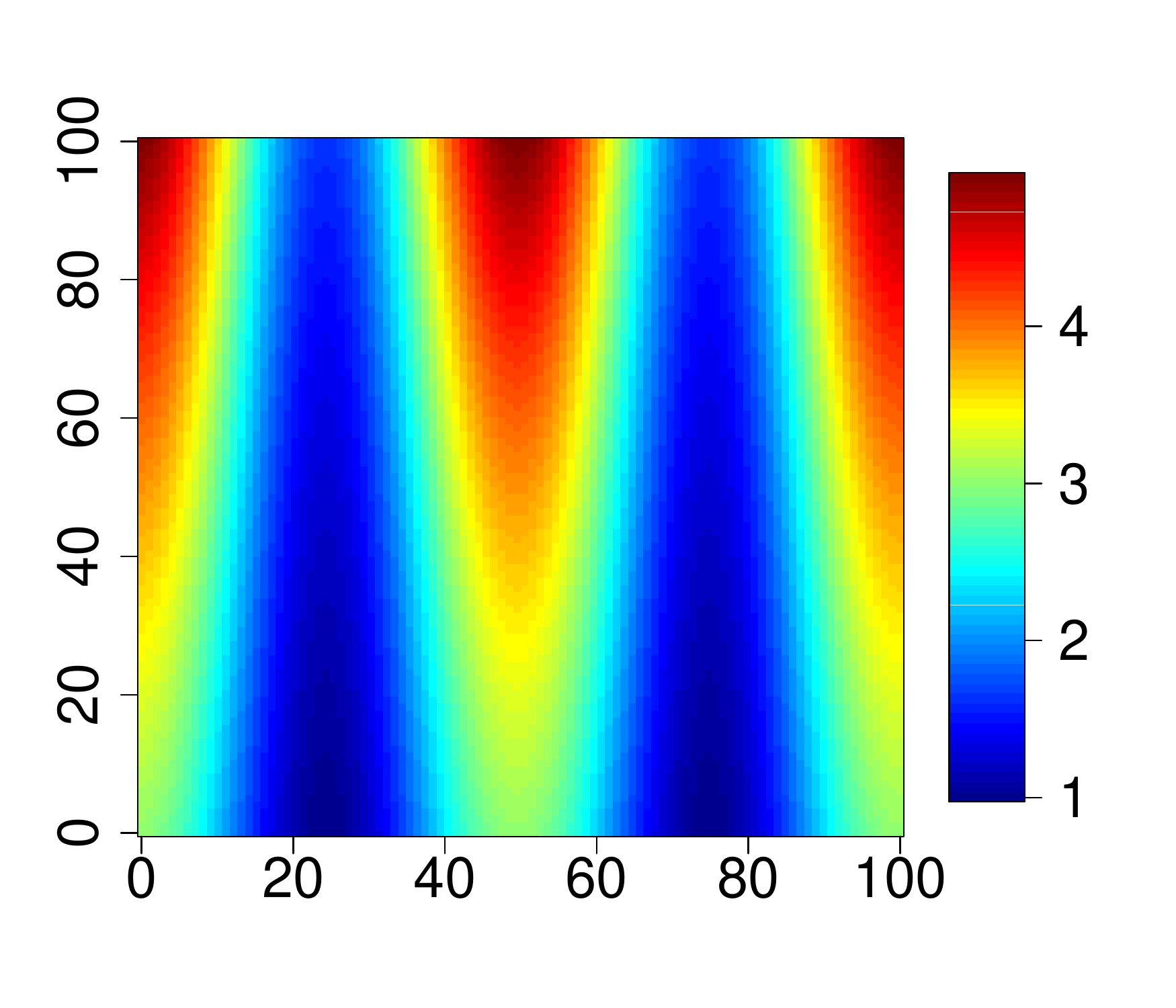}
			\subcaption{ 	\small Range parameters}
	\end{subfigure}
	\vline
	\begin{subfigure}[t]{.24\columnwidth}
		\centering\includegraphics[width=1\columnwidth]{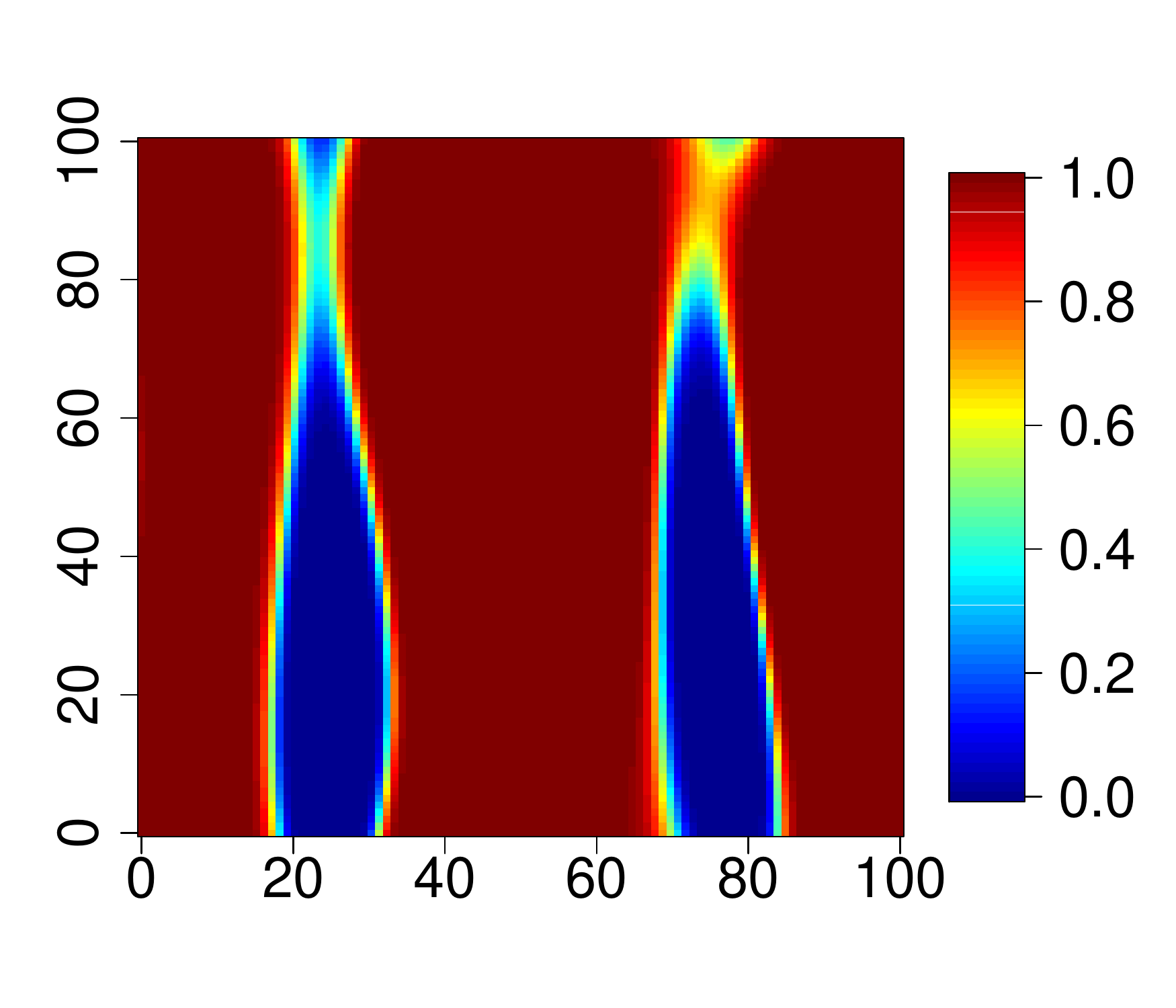}
		\subcaption{Weight field 1}
	\end{subfigure}
		\begin{subfigure}[t]{.24\columnwidth}
			\centering\includegraphics[width=1\columnwidth]{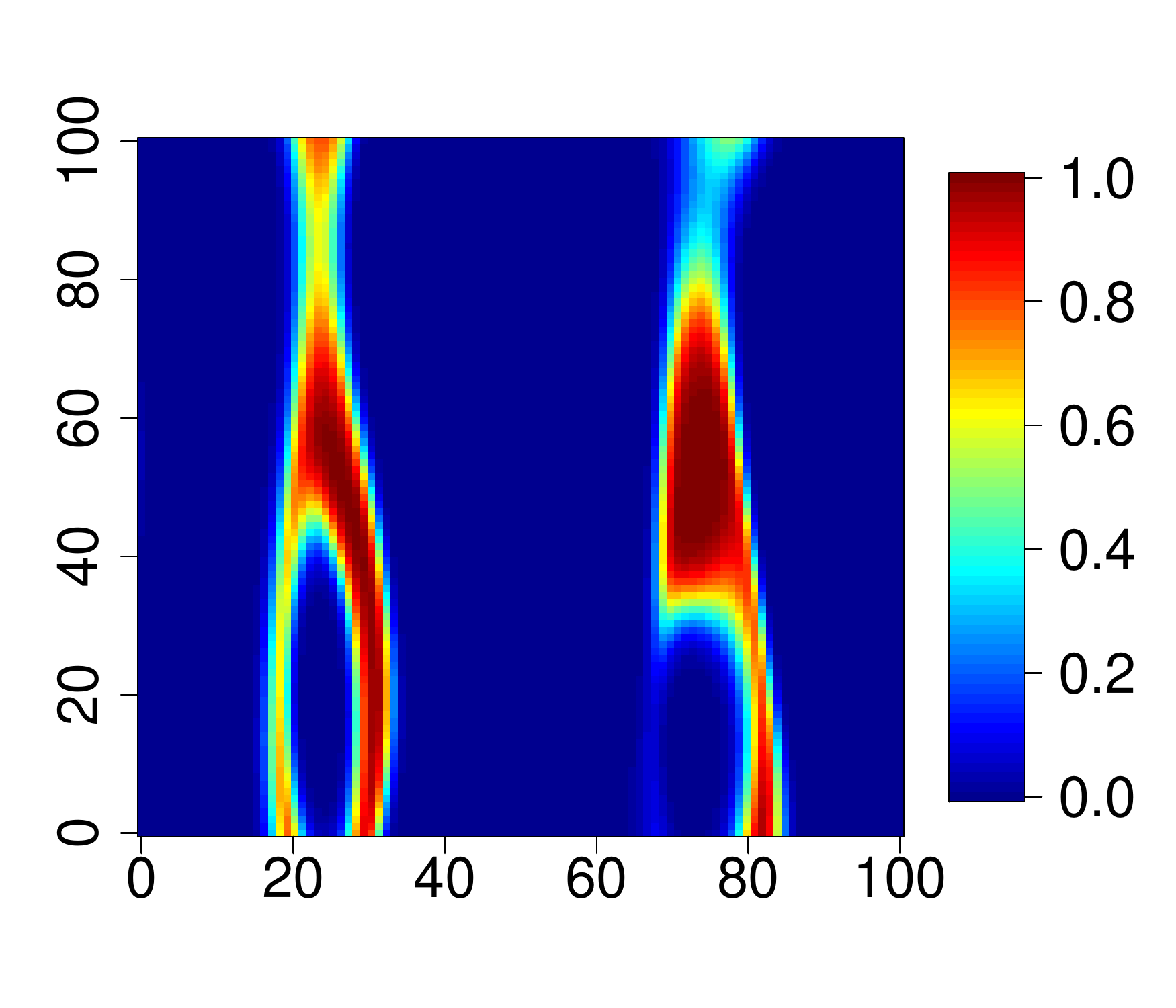}
			\subcaption{Weight field 2}
		\end{subfigure}
				\begin{subfigure}[t]{.24\columnwidth}
					\centering\includegraphics[width=1\columnwidth]{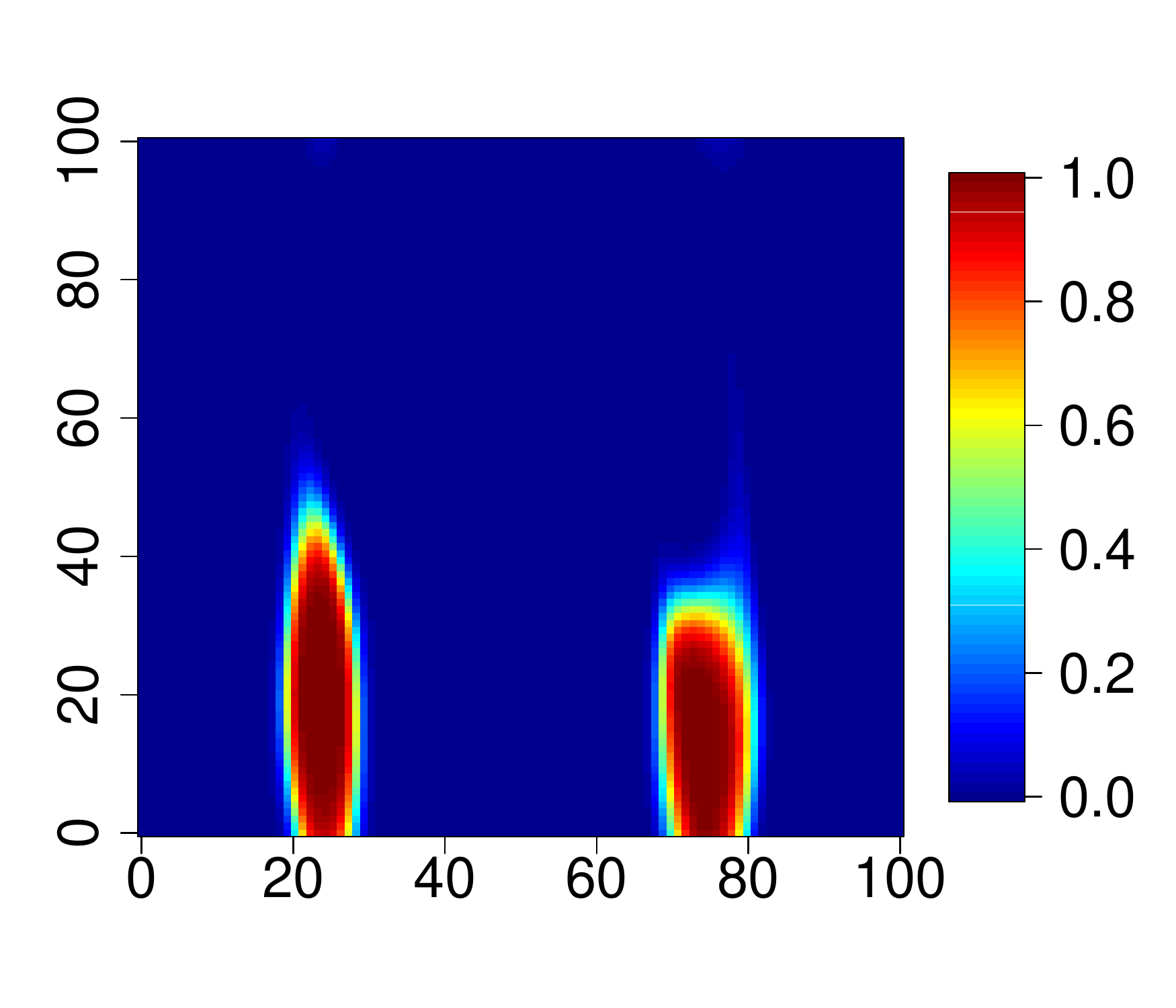}
					\subcaption{Weight field 3}
				\end{subfigure}
		\begin{subfigure}[t]{.24\columnwidth}
			\centering\includegraphics[width=1\columnwidth]{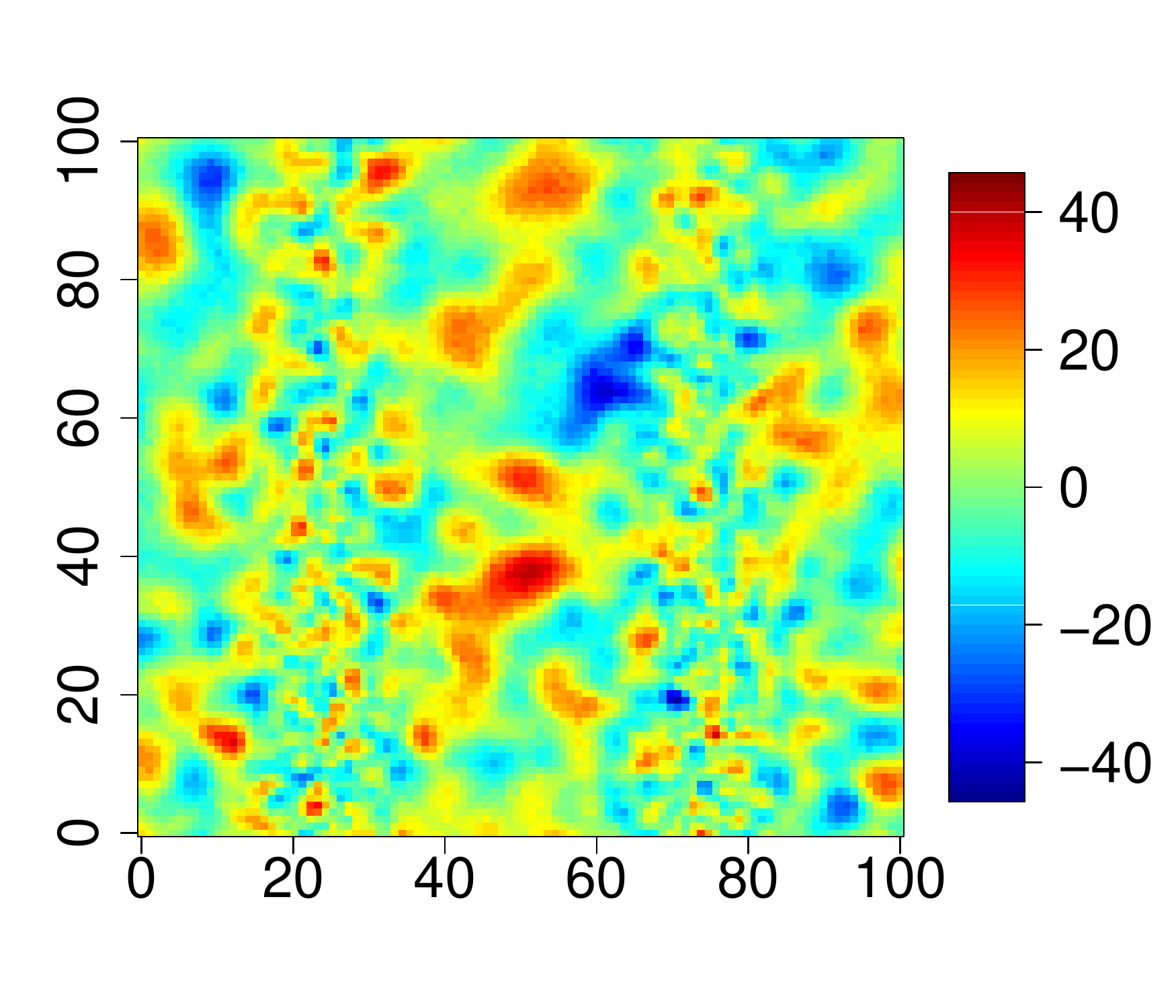}
			\subcaption{\small Synthesized Data}
		\end{subfigure}
			\vline
		\begin{subfigure}[t]{.24\columnwidth}
			\centering\includegraphics[width=1\columnwidth]{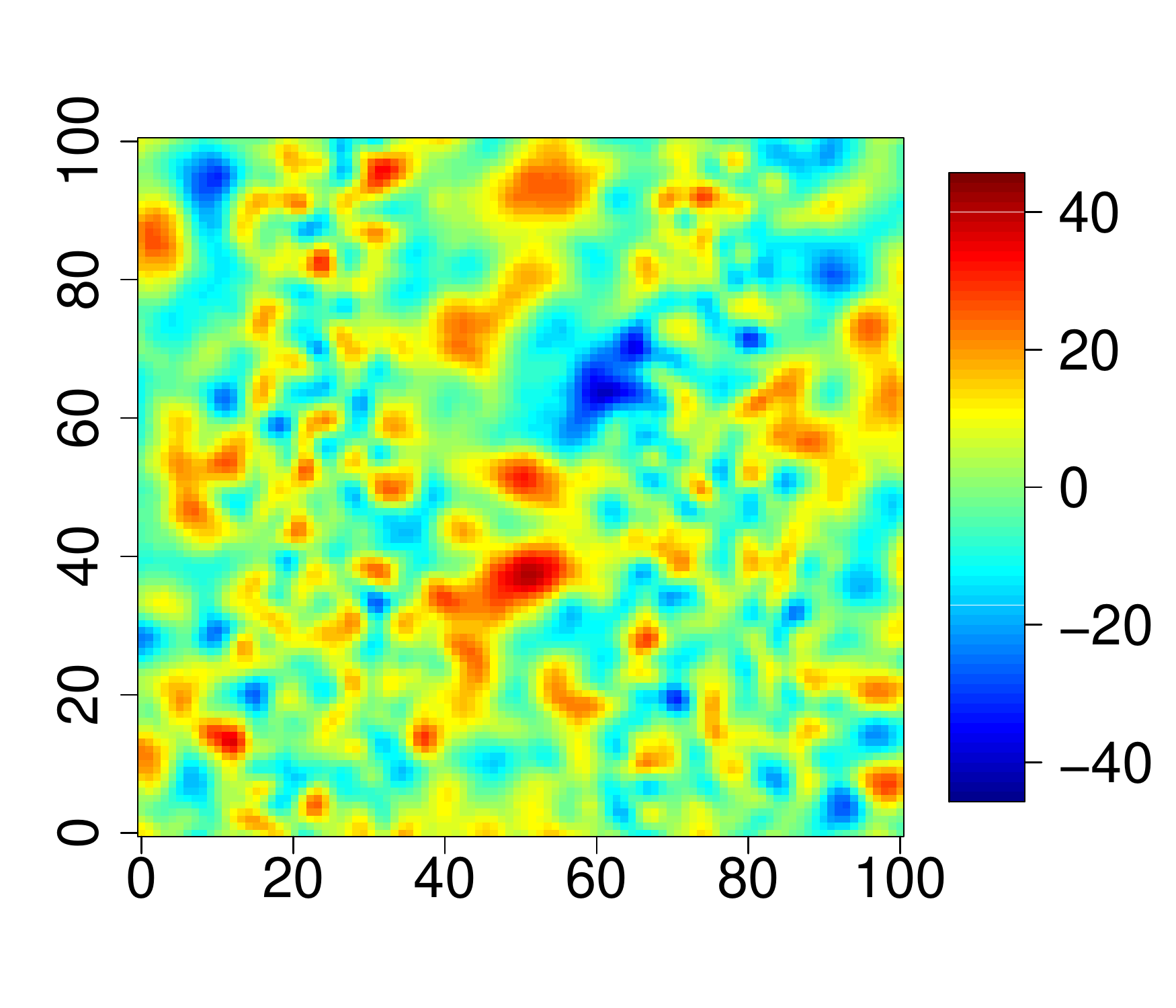}
			\subcaption{Mean field 1}
		\end{subfigure}
		\begin{subfigure}[t]{.24\columnwidth}
			\centering\includegraphics[width=1\columnwidth]{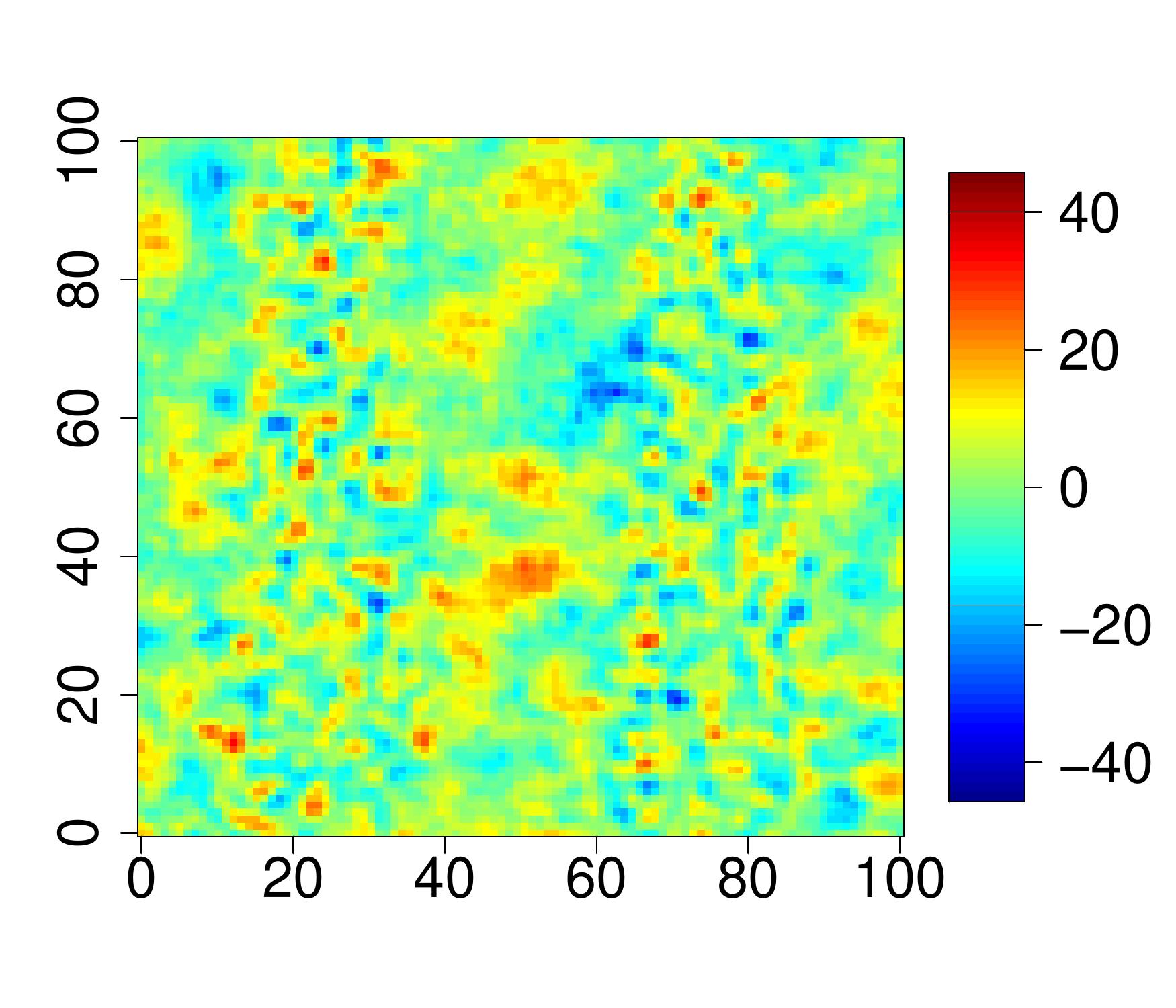}
			\subcaption{Mean field 2}
		\end{subfigure}
				\begin{subfigure}[t]{.24\columnwidth}
					\centering\includegraphics[width=1\columnwidth]{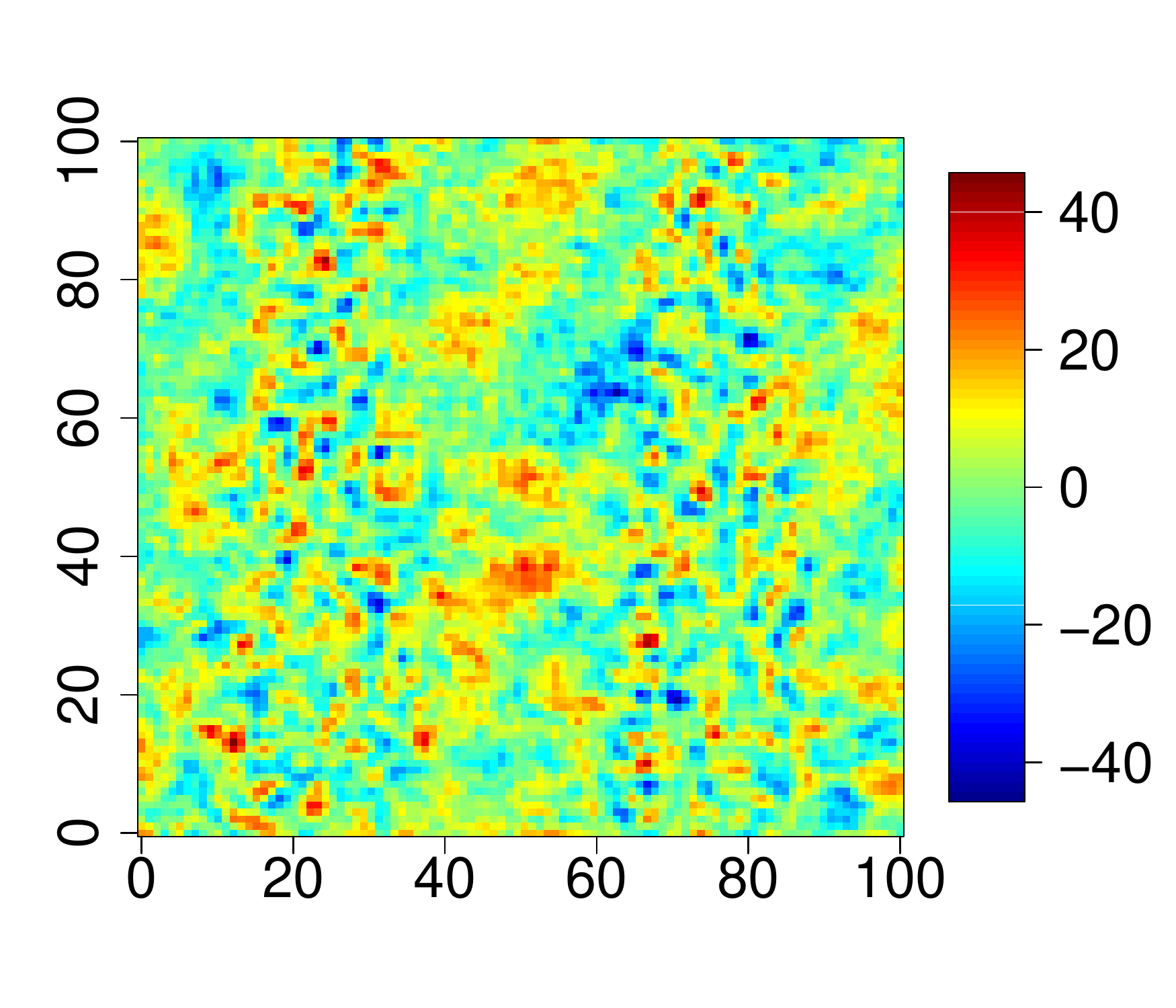}
					\subcaption{Mean field 3}
				\end{subfigure}
	\caption{Simulation for Non-Stationary FGP}
	\label{sim2}
\end{figure}

\begin{table}[ht]
	\centering
	\tiny
	\begin{tabular}{ l |l | l | l   }
		\hline
		 &	NS-FGP & 	S-FGP  & GPP (64 knots)  \\
		\hline		
		$\rho$ &  $2.78 (0.05)$, $1.55 (0.03)$ and  $1.20 (0.08)$ & 2.15 (0.15)& 3.56 (1.20)\\
		RMSE & $1.81$ & $4.96$ & $10.20$\\
		MAD & $1.62$ & $3.75$ & $6.96$\\
		 Time & 300 secs & 247 secs & 291 secs  \\
		\hline
	\end{tabular}
	\caption{Comparison of estimation for non-stationary Gaussian process}
	\label{sim_statioanry_comparison}
\end{table}

\subsection{Real Data Application}

We now apply the functional Gaussian process on a massive spatial-temporal dataset. The data are obtained from the North American Regional Climate Change Assessment Program (NARCCAP). We use the surface air temperature in the North America region \citep{mearns2011north}. The data are simulations from the Weather Research \& Forecasting regional model (WRF) coupled with the Third Generation Coupled Global Climate Model (CGCM3). We choose the daily average temperate in a 92-day period, from June 1st to August 31th in 2000. which has 1,343,752 data points. To evaluate the prediction performance, we randomly left out 20\% of the data as the testing set. This leads to training on 1,075,000 data points, which cost about $3$ GBs of computer memory for storage alone. Direct estimation of the fine scale matrix with size of 1,075,000$\times$1,075,000 would cost 3,225,000 GBs of memory, which is unrealistic for the modern computers. FGP provides a nice solution to this problem: since it uses the Fourier transform of the original matrix,
it involves at most the same size of the data for each stationary component; often, it cost less due to the sparsity of the spectral density (with sparse ratio commonly $<20\%$). We would like to emphasize that, unlike other approaches that uses lower-resolution grid for dimension reduction, FGP directly models the full scale matrix and does not involve resolution loss. For such a massive and fine-scale system, it took only 27 hours to finish each 30,000-step Markov-Chain Monte Carlo (MCMC) run.

We assume the observed temperature $\tilde{Z}$ at space/time point ${\bf s}=\{s_1,s_2,t\}$ is
$$\tilde{Z}_ {s_1,s_2,t}=\beta^T X_{s_1,s_2,t}+ Z_{s_1,s_2,t}+\epsilon_{s_1,s_2,t}$$
where ${s_1,s_2,t}$ represent the longitude, latitude and day, respectively; the term $X_{s_1,s_2,t}$ is a fixed linear term which contains the intercept, first and second order terms of $s_1,s_2,t$; the second term $Z_{s_1,s_2,t}$ is assumed be a location varying term; the last term $\epsilon(x_s,y_s,t_s)$ represents the random error, which is assumed $\epsilon_{s_1,s_2,t}\stackrel{indep}{\sim} N(0,\nu_{s_1,s_2,t}^2)$.

We applied this NS-FGP to model the spatial varying term $Z_{s_1,s_2,t}$.
We parameterize the non-stationary model, by assigning the component weight and component mean with the isotropic covariance function $Cov\{(s_1,s_2, t),(s_1',s'_2,t')\}= \phi exp(- \frac{|s_1-{s'}_1|^2}{2\rho_1^2} - \frac{|s_2-{s'}_2|^2}{2\rho_2^2} - \frac{|t-{t'}|^2}{2\rho_t^2})+\sigma^2 1_{(s_1,s_2, t)=(s_1',s'_2,t')}$. The spectral density is the product of three Fourier transforms. We also tested multiplying an extra space-time interaction term $exp(- \frac{|s_1-{s'}_1|^2|t-{t'}|^2}{c_1}- \frac{|s_2-{s'}_2|^2|t-{t'}|^2}{c_2})$ to the first term, as suggested by \cite{cressie1999classes}. Nevertheless, we found  that the posterior estimates of $c_1$ and $c_2$ for this data are quite large $(>10^5)$, so that the interaction term is negligible. Therefore, we restrict the following analyses on the non-interactive isotropic model.

We ran MCMC sampling for 30,000 steps and use the last 20,000 steps with 10-step thinning as the posterior sample. To approximate the infinite component assumption, we started with 16 components. The NS-FGP quickly converges to 3 major components. The results for parameter estimation are listed in Table~\ref{real_data_app}. We use the truncation rule of $g \ge 0.01 \sigma^2$ in likelihood evaluation, which leads a dimension reduction in the  spectral density values from 1,075,000 to only $\sim$53,000 ($\sim 5\%$ truncation rate). We repeated the sampling for 3 times using different random numbers as the starting values, and found the model converges to similar configurations and close parameter estimates (Figure~\ref{mcmc_diag}(a)). And the trace also suggests the model has good convergence (Figure~\ref{mcmc_diag}(b)).

\begin{figure}[ht]
	\centering
	\begin{subfigure}[t]{.4\columnwidth}
		\centering\includegraphics[width=1\columnwidth]{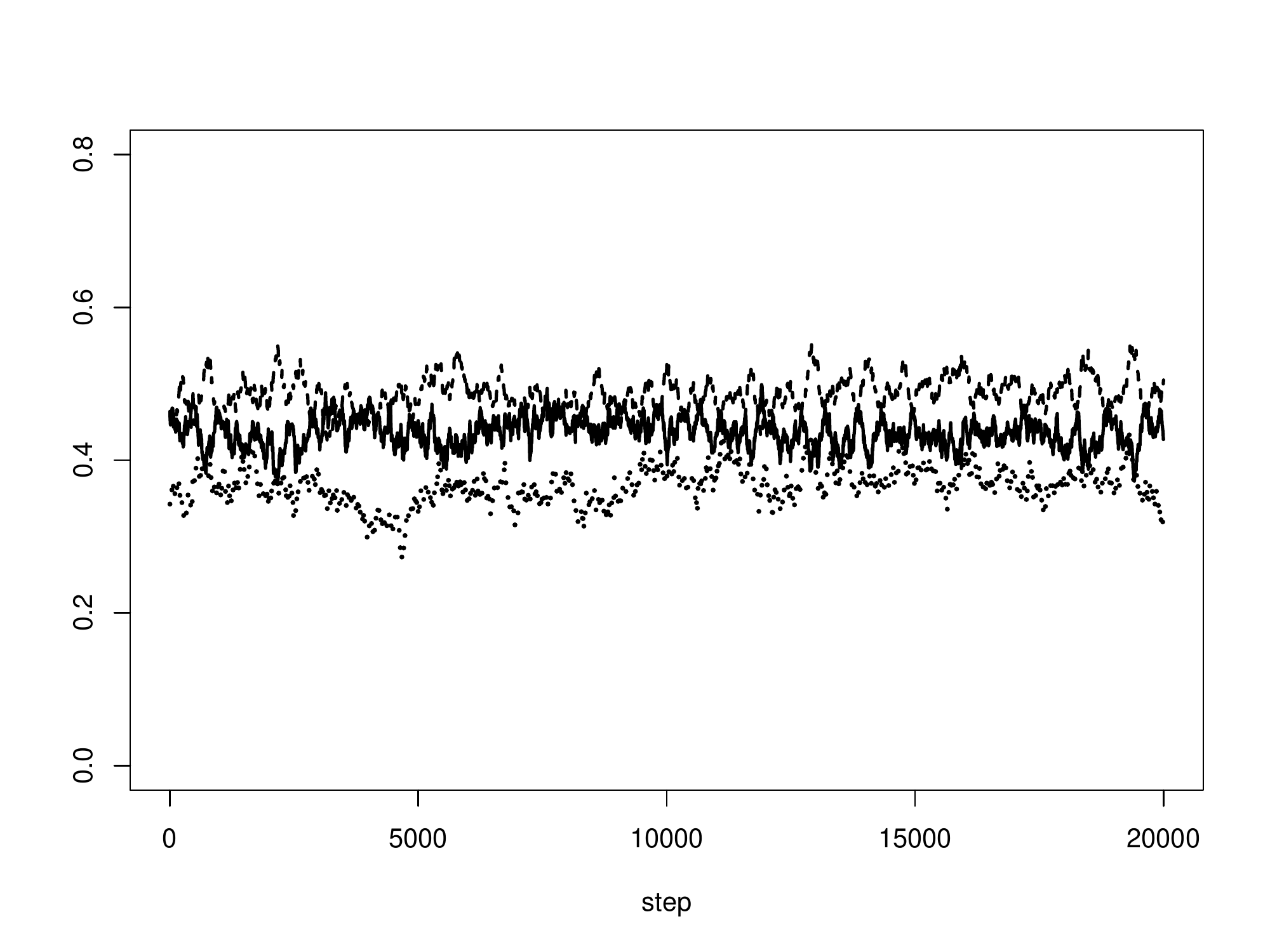}
		\subcaption{Trace plot of parameter $\rho_t$ for the 1st component mean, collected from 3 independet runs}
	\end{subfigure}	
		\begin{subfigure}[t]{.4\columnwidth}
			\centering\includegraphics[width=1\columnwidth]{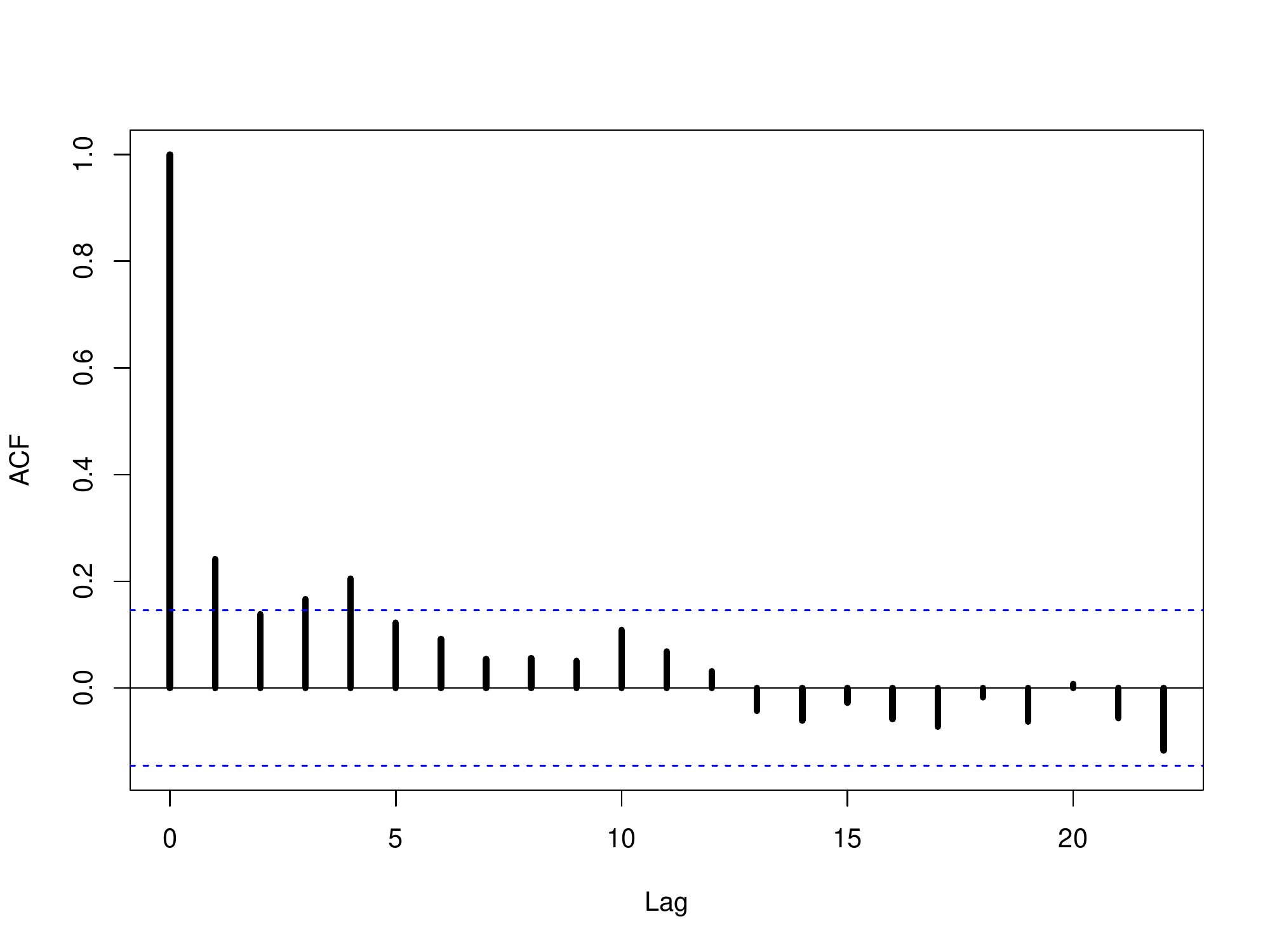}
			\subcaption{Autocorrelation plot of parameter $\rho_t$ for the 1st component mean}
		\end{subfigure}	
	\caption{The diagonistics show good mixing of the chain produced in the posterior sampling}
	\label{mcmc_diag}
\end{figure}

\begin{table}[ht]
	\scriptsize	
	\begin{center}
		\begin{tabular}{ r r  | r r r  | r}
						&& NS-FGP &	 & & Stationary FGP \\
			\hline                        
			&& Component 1&	Component 2 & Component 3  \\
			\hline
			\hline
			Mean &$\phi$ & 24.80  (1.52)& 38.27  (5.95) &9.89  (0.52) & 48.42 (5.62)\\
			&$\rho_1$ & 4.22 (0.05) & 1.67  (0.02)  & 2.51 (0.04) & 5.34 (0.45)\\
			&$\rho_2$  & 4.90 (0.03) & 2.19  (0.04) &3.05 (0.05) & 5.75 (0.56)\\
			&$\rho_t$  & 0.43  (0.02) & 0.09  (0.01) & 2.95  (0.12) & 0.54 (0.06)\\
%			$c_1$  & 6347 (451) & 61.72 (2.01) & 2400 (145)\\
%			$c_2$  &4615 (239) &   73.38 (1.21) &2396 (146) \\
			\hline
			Weight &$\phi$ & 1.58  (1.21)& 57.45  (15.16) & 65.05  (15.44)\\
			&$\rho_1$   & 20.13 (2.53) & 5.15  (0.11)  & 4.13 (0.09) \\
			&$\rho_2$  & 15.41 (1.20) & 6.76  (0.15) &5.31 (0.12) \\
			&$\rho_t$ & 12.04  (1.25) & 0.73  (0.02) & 0.67  (0.02)\\
%			$c_1$ for $\boldsymbol{\mu}_{zl}$ & 6347 (451) & 61.72 (2.01) & 2400 (145)\\
%			$c_2$ for $\boldsymbol{\mu}_{zl}$ &4615 (239) &   73.38 (1.21) &2396 (146) \\
						\hline
			\hline
			Clustering &$C$ Proportion ( \%) & 64\% & 17\% & 19\% \\
						\hline
			\hline
				Prediction Performance		& RMSE & 1.13 &&& 2.75\\
					& MAD & 0.65 &&& 1.26\\
						\hline
		\end{tabular}
	\end{center}
	\caption{The parameter estimates and cross-validation performance in non-stationary and stationary model.}
	\label{real_data_app}
\end{table}

We plot the data without the estimated trend $\beta^T X_{s_1,s_2,t}$ ( Figure~\ref{Fig:4_Components_NS} (a)), the mean estimate  (Figure~\ref{Fig:4_Components_NS} (b)) and the weight estimate  (Figure~\ref{Fig:4_Components_NS} (c)) for each major component from June 1st,2000 to June 3rd, 2000 . The plots are organized by columns and each column represents one day. The NS-FGP captures different strengths of correlation in the mean estimates. Compared vertically, the three components seem to correspond to the oceanic, coastal and the localized whether patterns in North America. Thanks to the large scale $\phi$ for the latent weight process $\bf L$, we see the weight $p$ close to 0 or 1 in most locations. This suggests a quite stable clustering, conditioning on which we can claim joint normality in the whole region. Since we used daily temperature data, we observed large variation of temperature pattern from day to day (compared horizontally in Figure~\ref{Fig:4_Components_NS} (a)). In the model estimates, we indeed saw low temporal correlation (Table~\ref{real_data_app}) and dynamic changes in the distribution of the mean and the weight (Figure~\ref{Fig:4_Components_NS} (b,c)).

\begin{figure}[hp]
	\centering
		\begin{subfigure}[t]{.7\columnwidth}
			\centering\includegraphics[width=1\columnwidth]{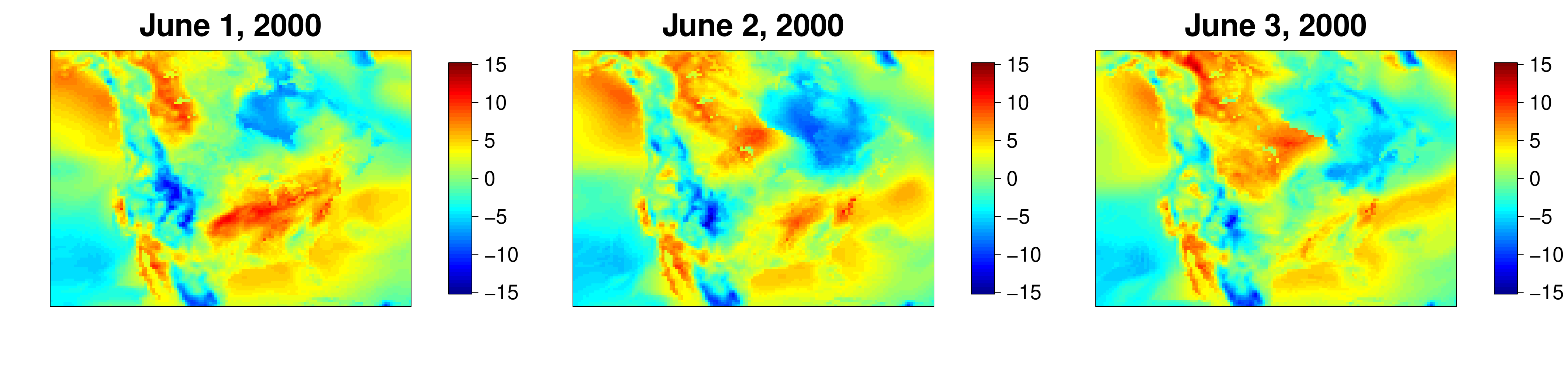}
			\subcaption{De-trended Data}
		\end{subfigure}	
	\begin{subfigure}[t]{.7\columnwidth}
		\centering\includegraphics[width=1\columnwidth]{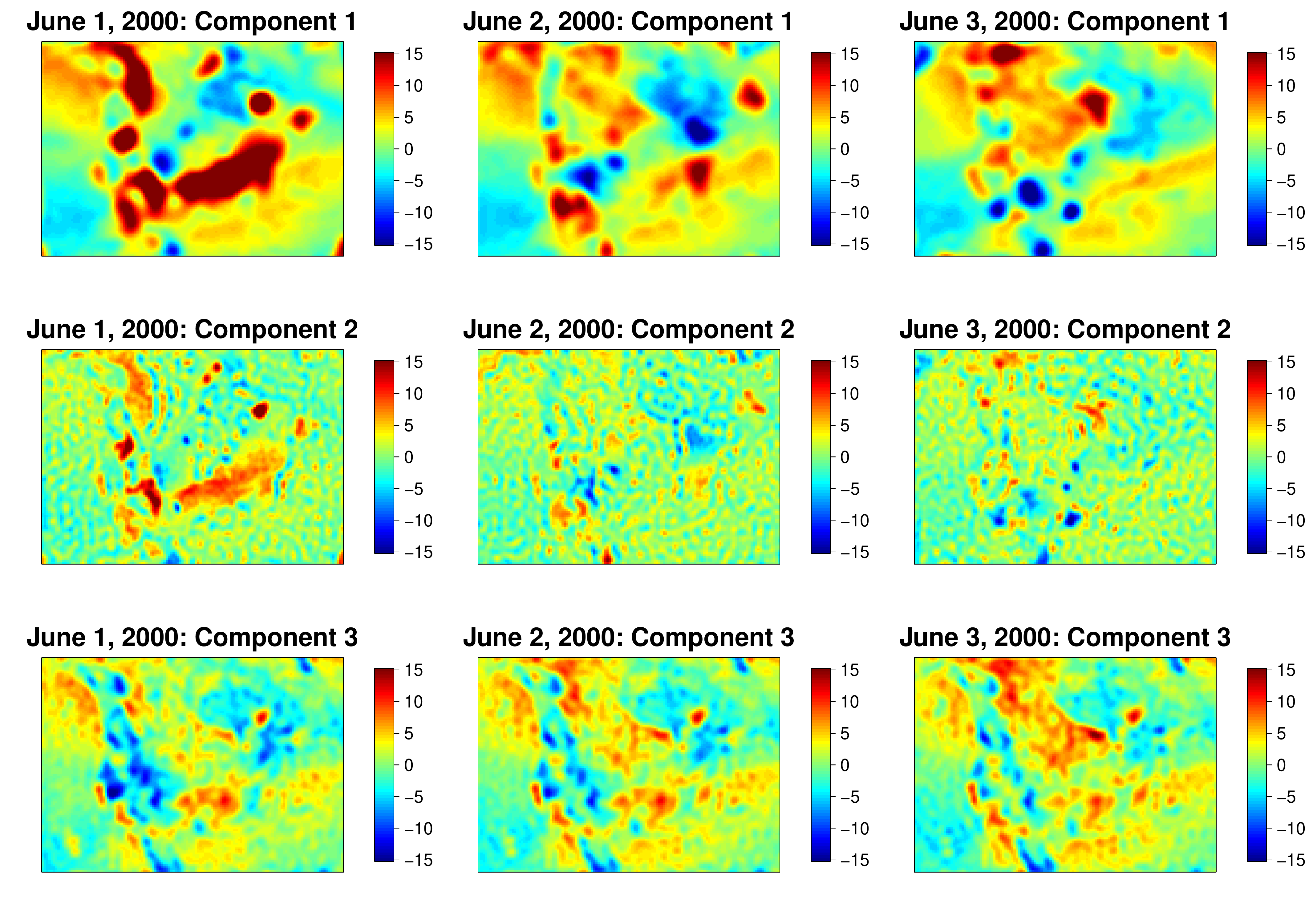}
		\subcaption{Component mean}
	\end{subfigure}	
		\begin{subfigure}[t]{.7\columnwidth}
			\centering\includegraphics[width=1\columnwidth]{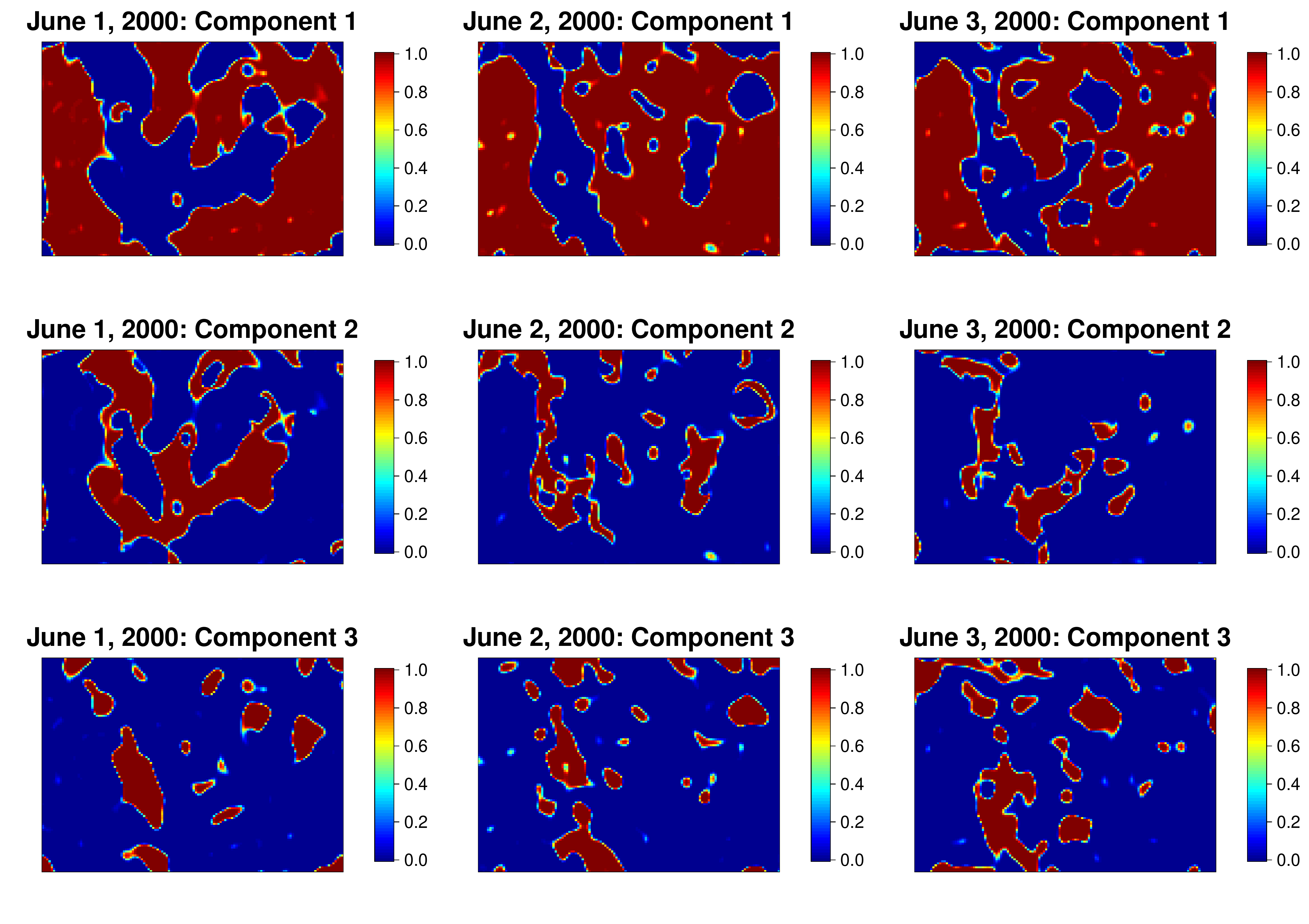}
			\subcaption{Component weight}
		\end{subfigure}	
	\caption{The surface temperature data in three days and the three stationary components estimated by NS-FGP}
	\label{Fig:4_Components_NS}
\end{figure}

Lastly we tested the prediction performance of the NS-FGP. Since we do not know the cluster assignment for the predicted locations, we first use the $\hat C_{{\bf s}_j}=arg\max_k  p_{{\bf s}_j, k}$ as the estimator for $C_{{\bf s}_j}$. As mentioned above, we have $\max_k  p_{{\bf s}_j, k}\approx 1$  in almost all the locations. In the prediction metrics, as shown in Table~\ref{real_data_app},  NS-FGP produced quite accurate prediction. As a comparison, we also ran the stationary FGP model on the data. The stationary FGP seemed to overly smooth the data, therefore is less accurate than the non-stationary model.

\section{Discussion and Future Work}

Our proposed method provides a new construction of Gaussian process that directly connects the spectral properties to its applications, such as parameter estimation and prediction. There are several extensions worth researching in the future. First, since space-time interaction is easier to obtain via spectral convolution than covariance function construction, it is interesting to relax the form of the spectral density, regardless of whether the closed form of covariance function exists. Second, in the non-stationary method, we provide a general  mixture framework with spectral dependency and we use probit stick-breaking process for illustration. Some other clustering approaches such as Pitman-Yor process \citep{ishwaran2001gibbs} may be studied to have more components. Third, the spectral properties in multivariate analysis can be studied, with a spatial correlation across different dependent variables. Fourth, more theoretic studies can be pursued, such as objective Bayesian priors and the posterior consistency.

\newpage
\begin{center}
{\large\bf SUPPLEMENTARY MATERIAL}
\end{center}

\subsection{Proof of Theorems}

\subsubsection{Proof of theorem 1}

We first prove the covariance (\ref{Eqn:covariance_function}) is real. Because of the symmetric function in $g(\{\omega_1,...,\omega_k,...,\omega_d\})=g(\{\omega_1,...,-\omega_k,...,\omega_d\})$ for any $k=1,2,...,d$ and symmetric distribution of $\{\boldsymbol{\omega}_l\}$ about $\bf 0$, we have
\begin{equation*}
\begin{aligned}
\sum_{l=1}^{n} exp\{i \boldsymbol{\omega}_l^T (\mathbf{s}_j-\mathbf{s}_k)\}g(\boldsymbol{\omega}_l) &= \sum_{l=1}^{n}\{\cos(\boldsymbol{\omega}_l^T (\mathbf{s}_j-\mathbf{s}_k)) g(\boldsymbol{\omega}_l)+ i \sin(\boldsymbol{\omega}_l^T (\mathbf{s}_j-\mathbf{s}_k))g(\boldsymbol{\omega}_l)\}\\
&= 2\sum_{l=1}^{\lfloor n/2\rfloor}\{\cos(\boldsymbol{\omega}_l^T (\mathbf{s}_j-\mathbf{s}_k)) g(\boldsymbol{\omega}_l)\} +g(\boldsymbol{0})
\end{aligned}	
\end{equation*}
where $n$ is assumed to be an odd number; if $n$ is even, we remove $g(\boldsymbol{0})$ on the right hand side. In either case, the imaginary parts are canceled.

To prove the positive definiteness, we use the matrix representation  $\boldsymbol{\Sigma} = {\bf Q} {\bf G}\mathbf{Q^*}+\mathbf{I}\sigma^2$. For any nontrivial real vector $\boldsymbol X$ of size $N$, we have:

\begin{equation*}
\begin{aligned}
\boldsymbol{X'\Sigma X} =\boldsymbol{X' {\bf Q} {\bf G}\mathbf{Q^*} X}+\boldsymbol{X'X}\sigma^2
\end{aligned}	
\end{equation*}

It is trivial that $\boldsymbol{X'X}\sigma^2>0$ for $\sigma^2>0$ . Now we denote $\mathbf{Q^*} \boldsymbol{X} = \boldsymbol{Y}= \boldsymbol{Y_1} +i\boldsymbol{Y_2} $, where $\boldsymbol{Y_1}$ and $\boldsymbol{Y_2}$ are the real and the imaginary parts of the transform of $\boldsymbol X$. We have $\boldsymbol{X' {\bf Q} {\bf G}\mathbf{Q^*} X} = \boldsymbol{Y_1}'{\bf G}\boldsymbol{Y_1} + \boldsymbol{Y_2}'{\bf G}\boldsymbol{Y_2}\ge 0$, as each element of $\bf G$ satisfies $g(.)\ge 0$. Combining two parts, we prove that for any nontrivial $\boldsymbol{X}$, $\boldsymbol{X'\Sigma X}>0$, which is the definition of positive definiteness.

\subsubsection{Proof of theorem 2}

As the covariance function without the nugget ${\sigma^2}$ can be viewed as $$C(\textbf{x})=\int_{\mathbb{R}^d} exp(i  \textbf{x}^T \boldsymbol\omega)\int_{\mathbb{R}^d} exp(-i  \textbf{x}^T \boldsymbol\omega) C(\textbf{x}) d{\bf x}d \boldsymbol\omega=\int_{\mathbb{R}^d} exp(i  \textbf{x}^T \boldsymbol\omega)g(\boldsymbol\omega)d\boldsymbol\omega$$.

Denote the specified covariance function as $Cov({\bf x})= Cov(\mathbf{s}_j,\mathbf{s}_k)$ and the subregion $\mathbb{W}=\{-\frac{m_1}{n_1}\Delta_1,-\frac{m_1-1}{n_1}\Delta_1,...,\frac{m_1}{n_1}\Delta_1\} \times ... \times \{-\frac{m_d}{n_d}\Delta_d,-\frac{m_d-1}{n_d}\Delta_d,...,\frac{m_d}{n_d}\Delta_d\}$ then we have:

$$ ||Cov({\bf x})-C({\bf x})|| \le ||\int_{\boldsymbol\omega \in \mathbb{W}} exp(i  \textbf{x}^T \boldsymbol\omega)g(\boldsymbol\omega)d\boldsymbol\omega-\sum_{l=1}^{m} exp\{i{\bf x}^T \boldsymbol{\omega}_l \}g(\boldsymbol{\omega}_l)/n||+  \int_{\boldsymbol\omega \not\in \mathbb{W}}\cos(\textbf{x}^T \boldsymbol\omega)g(\boldsymbol\omega)d\boldsymbol\omega$$

Since $||\cos(\textbf{x}^T \boldsymbol\omega)g(\boldsymbol\omega)|| \le \epsilon$, we let $\epsilon=1/m^2$ and use the dominated convergence theorem  $\lim_{m\rightarrow\infty}\int_{\boldsymbol\omega \not\in \mathbb{W}}\cos(\textbf{x}^T \boldsymbol\omega)g(\boldsymbol\omega)d\boldsymbol\omega = \int_{\boldsymbol\omega \not\in \mathbb{W}}\lim_{m\rightarrow\infty}\cos(\textbf{x}^T \boldsymbol\omega)g(\boldsymbol\omega)d\boldsymbol\omega=0$. And the first part corresponds to the error of the middle Riemann sum:

$$\lim_{m\rightarrow \infty}||\int_{\boldsymbol\omega \in \mathbb{W}} exp(i  \textbf{x}^T \boldsymbol\omega)g(\boldsymbol\omega)d\boldsymbol\omega-\sum_{l=1}^{m} exp\{i{\bf x}^T \boldsymbol{\omega}_l \}g(\boldsymbol{\omega}_l)/n|| \le K/m^2$$
where K is a finite constant.

\subsubsection{Proof of theorem 3}

We first prove the exchangeable condition. For any permutation of location vector ${\bf S}_\pi=\{\mathbf{s}_{\pi1},\mathbf{s}_{\pi2},...,\mathbf{s}_{\pi n}\}$ and the random variables $\mathbf{Z_{S_\pi}}$, we define the permutation matrix $\bf P_\pi$ such that $\bf S_\pi=P_\pi S $. Then we have $\mathbf{Z_{S_\pi}} =\bf P_\pi \mathbf{Z_{S}} $, $\mathbf{Q_{S_\pi}} =\bf P_\pi \mathbf{Q_{S}} $ and $\bf P_\pi \bf P'_\pi =\bf P'_\pi \bf P_\pi= I$.

Since $(\mathbf{P_\pi}\boldsymbol{\Sigma}\mathbf{P'_\pi})^{-1}=\mathbf{P_\pi}\boldsymbol{\Sigma}^{-1}\mathbf{P'_\pi}$ and $|\mathbf{P_\pi}\boldsymbol{\Sigma}\mathbf{P'_\pi}|=|\mathbf{P'_\pi}\mathbf{P_\pi}\boldsymbol{\Sigma}|=|\boldsymbol{\Sigma}|$ for any positive definite $\boldsymbol{\Sigma}$, which we showed for $\mathbf{Z_{S_\pi}}$ in theorem 1.

\begin{equation*}
\begin{aligned}
p(\mathbf{s}_{\pi1},\mathbf{s}_{\pi2},...,\mathbf{s}_{\pi n})
=&(2\pi)^{-n/2}|\mathbf{P_\pi}\boldsymbol{\Sigma}\mathbf{P'_\pi}|^{-1/2} exp(-\mathbf{Z'_{S}}\mathbf{P'_\pi}(\mathbf{P_\pi}\boldsymbol{\Sigma}\mathbf{P'_\pi})^{-1}\mathbf{P_\pi}\mathbf{Z_{S}}/2) \\
=&(2\pi)^{-n/2}|\boldsymbol{\Sigma}|^{-1/2} exp(-\mathbf{Z'_{S}} \boldsymbol{\Sigma}^{-1}\mathbf{Z_{S}}/2) \\
=& p(\mathbf{s}_{1},\mathbf{s}_{2},...,\mathbf{s}_{n})
\end{aligned}
\end{equation*}

Next, for a finite location set ${\bf S_0}=\{\mathbf{s}_{1},\mathbf{s}_{2},...,\mathbf{s}_{ n}\}$ and any location $\mathbf{s}_{k}\in \mathbb{R}^d$, we have the joint distribution:

\begin{equation*}
\begin{aligned}
\begin{bmatrix} {\bf Z_{\bf S_0}}\\ Z_{\mathbf{s}_{k}}
\end{bmatrix}
\sim N(
\begin{bmatrix}{\bf 0}\\0
\end{bmatrix},
\begin{bmatrix} 
{\mathbf {Q_{S}} }{\mathbf G}\mathbf{Q^*_{S}}+\mathbf{I}\sigma^2 &
{\mathbf {Q_{S}} }{\mathbf G}\mathbf{Q^*_{s_k}} \\
{\mathbf {Q_{s_k}} }{\mathbf G}\mathbf{Q^*_{S}} &
{\mathbf {Q_{s_k}} }{\mathbf G}\mathbf{Q^*_{s_k}}+\mathbf{I}\sigma^2
\end{bmatrix}
)
\end{aligned}
\end{equation*}

Using normal theory, it is straightforward to verify that:

	$$\int_{ \mathbb{R}^d} p(\mathbf{S_0},\mathbf{s}_{k}) d\mathbf{s}_{k}   =(2\pi)^{-n/2}|{\mathbf {Q_{S_0}} }{\mathbf G}\mathbf{Q^*_{S_0}}+\mathbf{I}\sigma^2|^{-1/2} exp(-\mathbf{Z'_{S_0}}({\mathbf {Q_{S_0}} }{\mathbf G}\mathbf{Q^*_{S_0}}+\mathbf{I}\sigma^2)^{-1}\mathbf{Z_{S_0}}/2) = 
	p(\mathbf{S_0})$$
	
\subsubsection{Proof of theorem 4}

The $j$th row of $\mathbf{Q^*_S}$, $\mathbf{Q^*_S}_j$, is an $(n_1n_2...n_d)$-element vector, and $j=j_1+...+j_d$ with $j_k\in \{ 0,...,m_k-1\}$ and $m_k \le n_k$. This row is composed of the $(n_1n_2...n_d)$ elements of the tensor product $q_{j_1}\otimes q_{j_2}\otimes...\otimes q_{j_d}$, where $q_{j_k}$ represents a vector in the $k$th sub-dimension:
$$q_{j_k} = \{ exp [-i  (-\frac{ n_k}{n_k}\pi)j_k ] /\sqrt{n_k}, exp[-i  (-\frac{ n_k-1}{n_k}\pi)j_k ]  /\sqrt{n_k},...,exp[ -i  (\frac{ n_k}{n_k}\pi)j_k] /\sqrt{n_k}\}   $$
which is the $j_k$th Fourier basis, which has the orthogonality. That is, given another row $l$,  if $j_k=l_k$ then $q'_{j_k}q^*_{l_k}=1$, else $q'_{j_k}q^*_{l_k}=0$. Using the rule of tensor product, we have $\mathbf{Q^*_S}_j \mathbf{Q_S}_l = 1$ only if $j=l$, else $0$.

When $m_k = n_k$, we have $\mathbf{Q'_S}=\mathbf{Q_S}$ and  $ \mathbf{Q_S} \mathbf{Q^*_S}= \mathbf{Q'_S}\mathbf{Q'^*_S}= (\mathbf{Q'^*_S})^*(\mathbf{Q'^*_S})$. Using the similar proof as above, except for changing $-i$ to $i$, we have $ \mathbf{Q_S} \mathbf{Q^*_S} = \bf I$.

\subsubsection{Proof of theorem 5}

We first show the positive definiteness of the non-stationary covariance function $ Cov ({Z_{\bf s_j}}, Z_{\bf s_k}| C_{\bf s_j},C_{\bf s_k})= {\mathbf {Q_{\bf s_j}} } (\mathbf G^{1/2}_{C_{\bf s_j}}\mathbf G^{1/2}_{C_{\bf s_k}})  \mathbf{Q^*_{\bf s_k}}+ \sigma^2 1_{j=k}$. For simplicity of notation, we abbreviate $\bf Q_{\bf s_j}$ as $\bf Q_{j}$ and $\mathbf G^{1/2}_{C_{\bf s_j}}$ as $\mathbf G^{1/2}_j$. Then we have the follow matrix decomposition:

\begin{equation*}
\begin{aligned}
&\begin{bmatrix} 
{\mathbf {Q_{1}} }{\mathbf G^{1/2}_1}{\mathbf G^{1/2}_1}\mathbf{Q^*_{1}}+\sigma^2 &
{\mathbf {Q_{1}} }{\mathbf G^{1/2}_1}{\mathbf G^{1/2}_2}\mathbf{Q^*_{2}} &
... &
{\mathbf {Q_{1}} }{\mathbf G^{1/2}_1}{\mathbf G^{1/2}_n}\mathbf{Q^*_{n}}
 \\
{\mathbf {Q_{2}} }{\mathbf G^{1/2}_2}{\mathbf G^{1/2}_1}\mathbf{Q^*_{1}} & ... &...&{\mathbf {Q_{2}} }{\mathbf G^{1/2}_2}{\mathbf G^{1/2}_n}\mathbf{Q^*_{n}} \\
...  & ... &...&...\\
{\mathbf {Q_{n}} }{\mathbf G^{1/2}_n}{\mathbf G^{1/2}_1}\mathbf{Q^*_{1}} & {\mathbf {Q_{n}} }{\mathbf G^{1/2}_n}{\mathbf G^{1/2}_2}\mathbf{Q^*_{2}}  &...&
{\mathbf {Q_{n}} }{\mathbf G^{1/2}_n}{\mathbf G^{1/2}_n}\mathbf{Q^*_{n}}+\sigma^2 
\end{bmatrix}
\\=&
\begin{bmatrix} 
\mathbf {Q_{1}} & \bf 0'  &...  & \bf 0' \\
\bf 0' & \mathbf {Q_{2}} &  ...  & \bf 0' \\
... & ...& ...&... \\
 \bf 0'  &\bf 0'  &  ...  &  \mathbf {Q_{n}} \\
\end{bmatrix}
\begin{bmatrix}
{\mathbf G^{1/2}_1} \\ {\mathbf G^{1/2}_2} \\...\\{\mathbf G^{1/2}_n}
\end{bmatrix}
\begin{bmatrix}
{\mathbf G^{1/2}_1} &  {\mathbf G^{1/2}_2} &...&{\mathbf G^{1/2}_n}
\end{bmatrix}
\begin{bmatrix} 
\mathbf {Q^*_{1}} & \bf 0  &...  & \bf 0 \\
\bf 0 & \mathbf {Q^*_{2}} &  ...  & \bf 0 \\
... & ...& ...&... \\
 \bf 0  &\bf 0  &  ...  &  \mathbf {Q^*_{n}} \\
\end{bmatrix}
+
\begin{bmatrix} 
\sigma^2&  0  &...  &  0 \\
 0 & \sigma^2 &  ...  &  0 \\
... & ...& ...&... \\
  0  & 0  &  ...  &  \sigma^2\\
\end{bmatrix}\\
=& \bf ABB'A^*+ I \sigma^2
\end{aligned}
\end{equation*}
where $\mathbf {Q_{(.)}}$ is $1$-by-$m$ matrix, $\bf 0$ is $n$-by-$1$ zero matrix and $\bf A$ and $\bf  B$ represent two corresponding block matrices.
For any nontrivial vector $\boldsymbol{X}$, we have  $\boldsymbol{X'\Sigma X} = {\boldsymbol X}' \bf ABB'A^* {\boldsymbol X}+  \boldsymbol {X'X} \sigma^2 = \boldsymbol {Y_1' Y_1} +\boldsymbol {Y'_2 Y_2}+  \boldsymbol {X'X} \sigma^2>0$, where ${\bf B'A}^* {\boldsymbol X}=\boldsymbol {Y_1}+i\boldsymbol {Y_2}$.

The two Kolmogorov consistency conditions are satisfied since the joint normal density is defined for every location set. The proof is similar to the proof of theorem 3.
\newpage

\bibliographystyle{chicago}

\bibliography{reference}
\end{document}